\documentclass[10pt,twocolumn,letterpaper]{article}

\usepackage{iccv}              

\usepackage{multirow}
\usepackage{multicol}
\usepackage{color}
\usepackage{makecell}
\usepackage{xcolor}
\usepackage{colortbl}

\def\para#1{\vspace{0.5em}\noindent\textbf{#1}}

\definecolor{iccvblue}{rgb}{0.21,0.49,0.74}
\usepackage[pagebackref,breaklinks,colorlinks,allcolors=iccvblue]{hyperref}

\def\paperID{4970} 
\def\confName{ICCV}
\def\confYear{2025}

\title{Rethink Sparse Signals for Pose-guided Text-to-image Generation}

\author{
    Wenjie~Xuan$^{1}$, Jing~Zhang$^{1}$, Juhua~Liu$^{1}$, Bo~Du$^{1}$, Dacheng~Tao$^{2}$\\
    \normalsize $^{1}$Wuhan~University, $^{2}$Nanyang~Technological~University
}

\begin{document}
    \twocolumn[{
        \renewcommand\twocolumn[1][]{#1}
        \maketitle
        \vspace{-3.0em}
        \begin{center}
            \centering
            \includegraphics[width=1.0\linewidth]{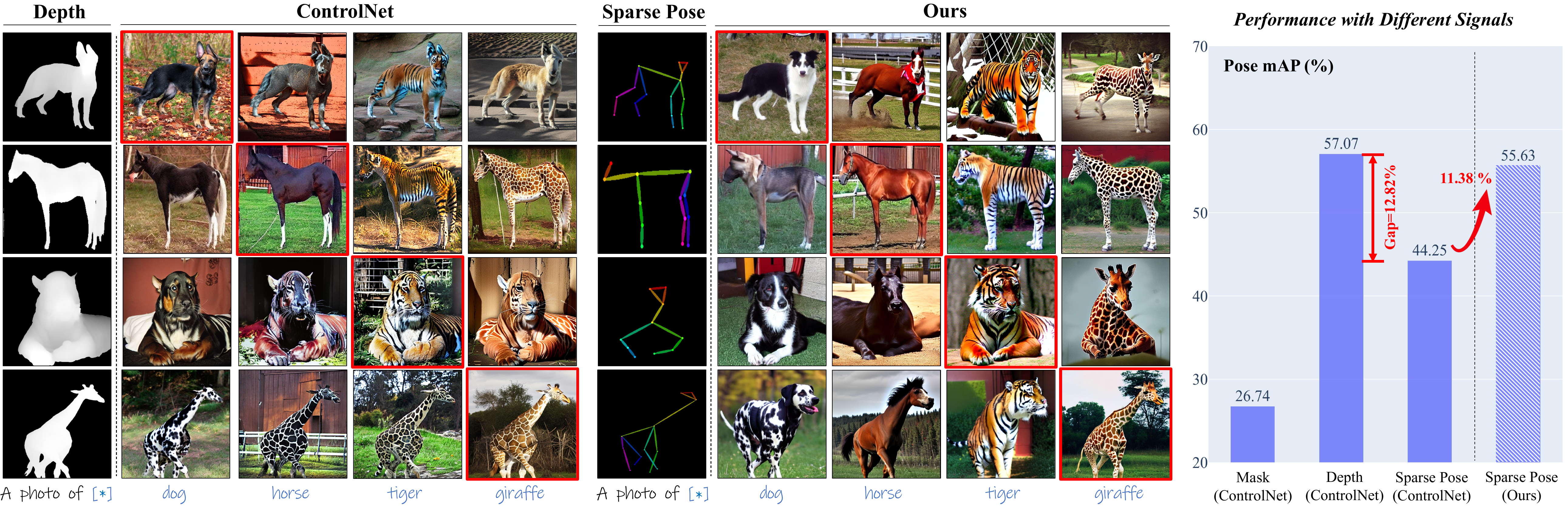}
            \vspace{-2.0em}
            \captionof{figure}{Performance of dense (\ie, depth) and sparse signals for pose-guided text-to-image generation. Though sparse pose signals are simple, they are supposed to suffer from poor pose control than dense signals. Our work improves the pose control of sparse signals, 
            achieving competitive pose alignment with dense signals while enjoying image diversity and cross-species generalization capability.
            }
            \label{fig:figure-1}
        \end{center}
    }]

\begin{abstract}
    Recent works favored dense signals (e.g., depth, DensePose), as an alternative to sparse signals (e.g., OpenPose), to provide detailed spatial guidance for pose-guided text-to-image generation. However, dense representations raised new challenges, including editing difficulties and potential inconsistencies with textual prompts. This fact motivates us to revisit sparse signals for pose guidance, owing to their simplicity and shape-agnostic nature, which remains underexplored. This paper proposes a novel {\bf S}patial-{\bf P}ose ControlNet~(SP-Ctrl), equipping sparse signals with robust controllability for pose-guided image generation. 
    Specifically, we extend OpenPose to a learnable spatial representation, making keypoint embeddings discriminative and expressive. Additionally, we introduce keypoint concept learning, which encourages keypoint tokens to attend to the spatial positions of each keypoint, thus improving pose alignment.
    Experiments on animal- and human-centric image generation tasks demonstrate that our method outperforms recent spatially controllable T2I generation approaches under sparse-pose guidance and even matches the performance of dense signal-based methods. 
    Moreover, SP-Ctrl shows promising capabilities in diverse and cross-species generation through sparse signals.
    Codes are available \href{https://github.com/DREAMXFAR/SP-Ctrl}{here}.
\end{abstract}

\section{Introduction}
\label{sec:intro}
    The recent success of pose-guided text-to-image~(T2I) generation~\cite{zhang2023adding,mou2024t2i,zhao2024uni,qin2023unicontrol,li2024controlnet++,peng2024controlnext,qiao2019mirrorgan} has encouraged applications including animal- and human-centric image generation~\cite{xu2024animatezoo,jiang2023spac,Ju_humansd,li2024ecnet,yin2024grpose}, pose-driven image animation~\cite{hu2024animate,xu2024magicanimate,zhu2024champ,chang2024magicpose}, and conditional 3D generation~\cite{patashnik2024consolidating,li2023mvcontrol}. Recent efforts tend to employ dense conditions ({\it e.g.}, depth and normal maps, DensePose, SMPL)~\cite{karras2023dreampose,liu2023hyperhuman,xu2024magicanimate,li2024dispose} or richer pose representations ({\it e.g.}, textual prompts, graphs, ViT features)~\cite{li2024ecnet,yin2024grpose,stablepose2024} for more precise pose control. However, as validated in Fig.~\ref{fig:figure-1}, such dense signals raised new challenges: 1) {\bf Inflexibility.} The dense pose signals are commonly extracted from reference images with pretrained detectors, not only restricting the creation of new conditions but also raising the difficulties for editing. 2) {\bf Contradictions with textual prompts.} Dense signals impose strong constraints on objects' shapes and contours. If these priors are violated, it would severely degrade fidelity and produce undesired results~\cite{qiao2019learn,xuan2024whencontrolnet,liu2024smartcontrol,jo2024skip}.  

    In contrast, sparse signals like OpenPose~\cite{openpose} inherently mitigate the above risks due to three appealing properties: 1) {\bf Shape-agnostic.} The sparse pose signal is an abstract of objects ({\it e.g.}, humans, animals) in anatomy. It is usually presented in graph structures, comprising a series of pre-defined keypoints and skeletons with topological structures. 2) {\bf Category-agnostic.} The definitions of keypoints are commonly shared in natural language, especially for animals. The topology structures are also similar among species like mammals despite subtle variances in skeleton proportions and actions, which provide a unified pose representation~\cite{yu2021ap}. 3) {\bf Operability.} The sparse signals do not necessarily rely on reference images, offering high freedom in creation and manipulation. However, though early methods like ControlNet~\cite{zhang2023adding} and T2I-Adapter~\cite{mou2024t2i} adopt OpenPose for spatial pose guidance, the sparsity of sparse signals raised great challenges for precise pose control, leading to poor spatial alignment with the given pose. Thus, a critical question raises: {\it Is it possible to achieve precise and robust pose-controllable generation using sparse signals?}

    This paper investigates sparse signals, {\it i.e.,} OpenPose, for pose-guided T2I generation and identifies two major bottlenecks for precise pose control. 
    First, since the OpenPose representation was originally designed for visualization purposes, its effectiveness for spatial pose guidance remains unclear. As shown in Fig.~\ref{fig:spe}, although OpenPose encodes keypoint locations and topological skeletons to the skeleton image, its color channels provide little information, which would even confuse the perception of distinctive keypoints as discussed in \S\ref{sec:ablation}.
    Second, the sparsity of pose signals, where keypoints and skeletons are usually denoted as vertices and edges separately, raises challenges for models to perceive and follow point-like spatial instructions~\cite{Li_2023_CVPR_gligen}. 
    Therefore, extra efforts are required to enhance the focus on keypoint locations and semantics. 
    
    In light of this, this paper proposes a spatial-pose ControlNet~({\it SP-Ctrl}), making it possible for precise pose-guided T2I generation through sparse signals. Specifically, we propose an expressive Spatial-Pose Representation~(SPR), which extends the plain RGB keypoint embeddings in OpenPose to learnable embeddings. Besides, to encourage pose alignment, we design a novel Keypoint Concept Learning~(KCL) strategy inspired by the associations between textual prompts and cross-attention maps~\cite{tang-etal-2023-daam}. We introduce new textual tokens to learn keypoint concepts through textual-inversion~\cite{gal2022textualinversion}, where a heatmap constraint is derived to encourage attention to each distinct keypoint for spatial alignment. Combining both strategies, our method largely increases the pose alignment of sparse-pose-guided T2I generation with little additional computational cost for inference. 
    Our method realizes performance comparable to dense signals and exhibits superior image diversity and cross-species generalization capability.

    Our contributions are summarized as follows. 
    \begin{enumerate}[leftmargin=0.8cm]
        \item[1)] We rethink the sparse signals for pose-guided T2I generation and propose {\it SP-Ctrl}, a spatial-pose ControlNet that enables precise pose alignment with sparse signals. It reveals the potential of sparse signals in spatially controllable generations. 
        \item[2)] We introduce a Spatial-Pose Representation with learnable keypoint embeddings to enhance the expressiveness of sparse pose signals.
        \item[3)] We propose Keypoint Concept Learning, a novel strategy that enhances keypoint focus and discrimination, enhancing details and improving pose alignment.
        \item[4)] Experiments on animal- and human-centric T2I generation validate that our method achieves performance comparable to dense signals. Moreover, our method advances in diversity and cross-species generation. 
    \end{enumerate}


    


    \begin{figure*}[t]
        \centering
        \includegraphics[width=0.98\linewidth]{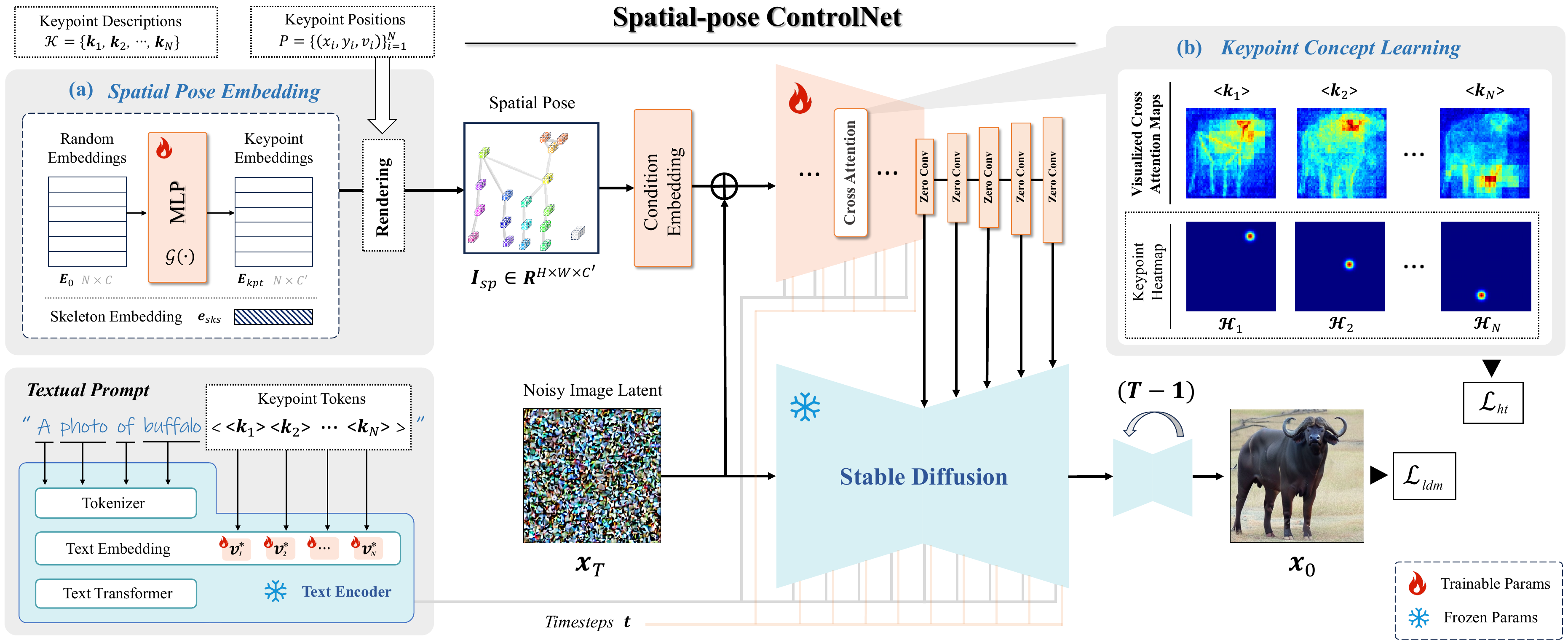}
        \caption{The overall architecture of our spatial-pose ControlNet~({\it SP-Ctrl}), mainly comprising: 1) a spatial-pose embedding module $\mathcal{G}(\cdot)$ to provide expressive sparse pose representations $I_{sp}$, 2) a set of new tokens $\{\boldsymbol{k}_i\}_{i=1}^N$ to learn the keypoint concept guided by heatmap constraints $\mathcal{L}_{ht}$ on the cross-attention maps, 3) a frozen SD model with a ControlNet adapter as a basic spatial controllable diffusion model. Our proposed model can synthesize images in high fidelity, which align precisely with the given sparse pose signals.}
        \vspace{-0.5em}
        \label{fig:arch}
    \end{figure*}

\section{Related Work}
\label{sec:related_work}
    \paragraph{Spatial Controllable T2I Diffusion Models.} With the popularity of text-to-image diffusion models like Stable Diffusion(SD)~\cite{Rombach_2022_CVPR_ldm}, recent works explored multi-modal image guidance ({\it e.g.}, edges, depths, pose, {\it etc.}) besides textual prompts to achieve spatially controllable synthesis. Early attempts~\cite{gafni2022make,avrahami2023spatext,huang2023composer} trained models from scratch with spatially aligned image-condition pairs that extracted by pretrained detectors. Another category of works~\cite{zhang2023adding,mou2024t2i,qin2023unicontrol,li2024controlnet++,zhao2024uni} introduced adapters to inject spatial control into the pre-trained T2I models, making it plug-and-play and composable for different conditions. Some efforts also explore training-free approaches~\cite{mo2024freecontrol,lin2024ctrlx}. Among these works, ControlNet~\cite{zhang2023adding} stands out for its exceptional spatial controllability and high fidelity, and its structure has been widely adopted in downstream applications, particularly for human image generation~\cite{Ju_humansd,cheong2023upgpt} and animation~\cite{xu2024magicanimate,hu2024animate,xu2024animatezoo}. This paper aims to enhance ControlNet's capability of interpreting sparse pose signals, enabling diverse outputs and cross-species generation with precise pose alignment.  

    \paragraph{Guidance for Pose Controllable Generation.} As an anatomical abstraction of objects with positions and definitions, pose signals can be provided in various formats to guide spatially controllable T2I generation. Early GAN-based methods~\cite{Men_2020_CVPR,zhang2021pise} and conditional diffusion models~\cite{bhunia2023person,shenadvancing,zhang2023adding,mou2024t2i} focused on the popular sparse OpenPose signals to drive image synthesis. However, recent studies revealed the limited guidance of sparse signals and adopted dense representations such as DensePose~\cite{karras2023dreampose,han2023controllable}, parsing masks~\cite{zhou2022cross,cheong2023upgpt}, and SMPL~\cite{liu2023hyperhuman}. While offering more precise control, these methods increased training complexity and limited the flexibility of creating and editing conditions. Meanwhile, a few works tended to enhance the controllability of sparse pose signals. For instance, HumanSD~\cite{Ju_humansd} introduced a perceptual loss with pre-trained pose estimators to improve alignment. Subsequent works explored strengthening sparse signals via textual prompts~\cite{li2024ecnet}, graph learning~\cite{yin2024grpose}, and ViT features~\cite{stablepose2024}. In contrast, this paper enriches sparse pose representations and introduces a novel keypoint concept learning strategy, realizing pose alignment without relying on pre-trained pose estimators. 

\section{Spatial-Pose ControlNet}
\label{sec:method}
    Given a sparse pose condition $\boldsymbol{c}_{p}$ and a textual prompt $\boldsymbol{c}_{t}$, our Spatial-pose ControlNet~({\it SP-Ctrl}) aims to synthesize images following the prompt with aligned poses. We first reveal two core designs, {\it i.e.}, Spatial-Pose Representation (\S\ref{sec:spe}) and Keypoint Concept Learning (\S\ref{sec:kcl}), followed by an introduction of the overall pipeline (\S\ref{sec:overall}). The architecture of our proposed {\it SP-Ctrl} is illustrated in Fig.~\ref{fig:arch}. 
    
    \subsection{Spatial-pose Representation}
    \label{sec:spe}
    
\begin{figure}[t]
            \centering
            \includegraphics[width=\linewidth]{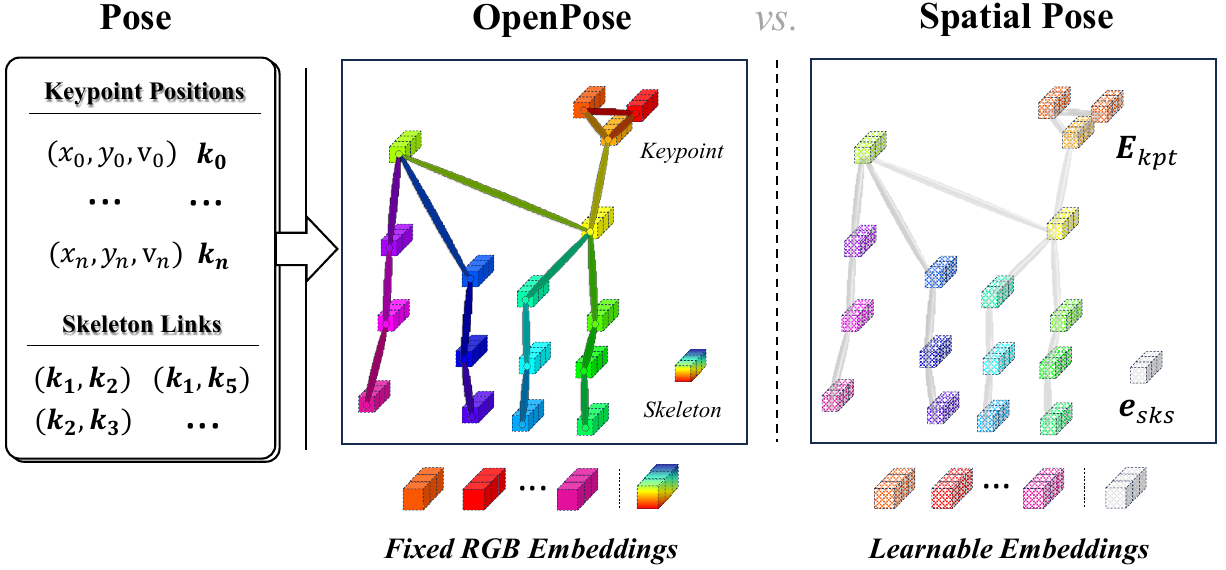}
            \caption{An illustration of OpenPose and our spatial pose.}
            \vspace{-1.0em}
            \label{fig:spe}
        \end{figure}

        Since OpenPose~\cite{openpose} was originally designed for visualization purposes, its functions for pose guidance have not been fully examined yet. Therefore, we first revisit OpenPose in the context of pose-guided generations. 
        Given $N$ pre-defined keypoints $P=\{(x_i, y_i, v_i)\}_{i=1}^{N}$, $(x_i, y_i)$ represents spatial coordinates and $v_i\in{\{0,1,2\}}$ indicates visibility, where $v_i\geq 1$ means the keypoint exists and is labeled in the image. 
        The keypoint descriptions are denoted as $\mathcal{K}=\{\boldsymbol{k}_i\}_{i=1}^{N}$, {\it e.g.}, eye, nose, elbow, etc. (See Appendix \S B for examples) The OpenPose renders skeleton pose images with uniformly sampled RGB colors as Fig.~\ref{fig:spe}, where vertices for keypoints and edges for skeletons. Though it is suitable for visual clarity, this representation ignores the semantic priors of $\mathcal{K}$, causing limited spatial guidance. 

        To this end, we develop a spatial pose representation~(SPR) for sparse-pose-guided generations. While inheriting the setting of OpenPose to render skeleton pose images indicating keypoint positions and skeletons, we extend the plain RGB embeddings of keypoints and skeletons to learnable embeddings inspired by SpaText~\cite{avrahami2023spatext}. This helps to encode more valuable insights into the sparse pose signals. Specifically, we employ a spatial-pose embedding module $\mathcal{G}(\cdot;\phi)$, which accepts random initialized vectors $\boldsymbol{E}_0=\{\boldsymbol{e}_k\in \mathbb{R}^{1\times C}\}_{k=1}^{N}$ as the inputs and produces embeddings for each keypoint, denoted as $\boldsymbol{E}_{kpt}=\{\boldsymbol{e}_k\in \mathbb{R}^{1\times C^\prime}\}_{k=1}^{N}$. This process is expressed as, 
        \begin{equation}
            \label{eq:2-1}
            \boldsymbol{E}_{kpt} = \mathcal{G}(\boldsymbol{E}_0;\phi), 
        \end{equation}
        where the spatial-pose embedding module $\mathcal{G}(\cdot)$ is an MLP parameterized by $\phi$ with stacked linear layers and layer norm. The detailed structure is presented in Appendix \S A. For skeleton embeddings, we use all-ones embeddings, denoted as $\boldsymbol{e}_{sks}=\mathbf{1}^{1\times C^\prime}$, which is proved to work well as discussed in \S\ref{sec:ablation}. Then, we render our spatial-pose representations with the learned embeddings $\boldsymbol{E}_{kpt}$ and $\boldsymbol{e}_{sks}$, resulting in a multi-channel skeleton pose image $\boldsymbol{I}_{sp}\in \mathbb{R}^{H\times W\times C^\prime}$ for pose guidance. Note that the spatial-pose embedding module $\mathcal{G}(\cdot)$ is optimized directly for the training target of denoising diffusion models, thus it learns spatially expressive keypoint embeddings $\boldsymbol{E}_{kpt}$ adaptively. Validations in \S\ref{sec:discussion} prove that the learned $\boldsymbol{E}_{kpt}$ effectively encodes semantic priors, where semantic-similar keypoints are pulled together, thus providing richer insight for pose guidance. 

        

    \subsection{Keypoint Concept Learning}
    \label{sec:kcl}

        \begin{figure}[t]
            \centering
            \includegraphics[width=0.96\linewidth]{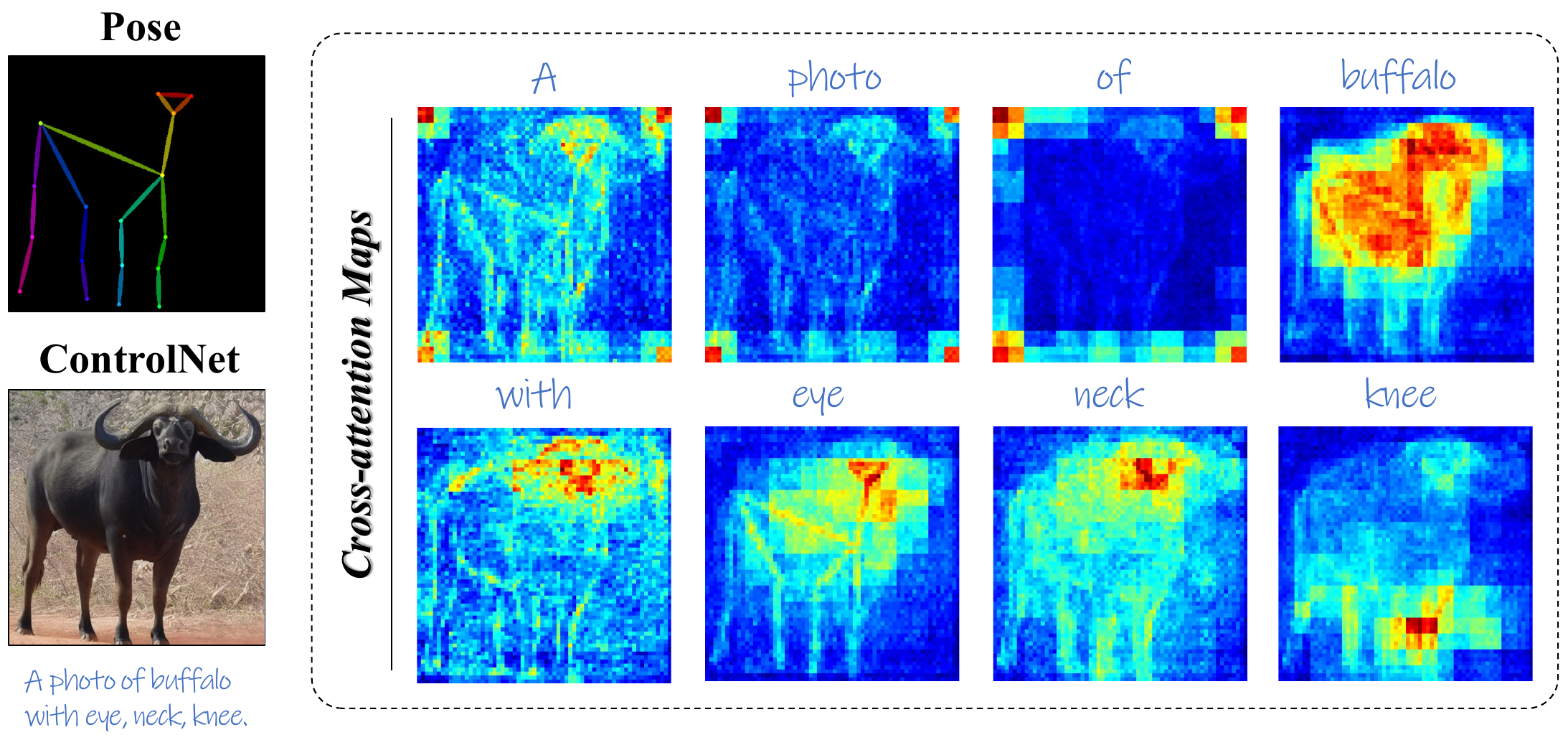}
            \caption{Visualized cross-attention maps of ControlNet at time step $799$, where the model attends to positions according to keypoint description. Full results refer to Appendix Fig.~S6.}
            \vspace{-1em}
            \label{fig:cross-attn-vis}
        \end{figure}
        
        Assuming that the pose alignment relies on both spatial perception and semantic understanding of keypoints, we analyze cross-attention maps to investigate the relevance between keypoint textual prompts and spatial responses in spatially controllable diffusion models, \ie, ControlNet. As illustrated in Fig.~\ref{fig:cross-attn-vis},  consistent with findings of prior studies~\cite{tang-etal-2023-daam,hertz2022prompttoprompt, Kim_2023denset2iedit}, the cross-attention maps exhibit strong correlations between nouns ({\it e.g.}, \texttt{buffalo}) and pixel positions. 
        Furthermore, the spatial responses of certain keypoint descriptions ({\it e.g.}, \texttt{eye}, \texttt{knee}) coarsely align with keypoint positions in OpenPose, especially at early denoising steps. This observation suggests that it is promising to improve pose alignment via strengthening the exact connection between spatial attention maps and each keypoint description.

        Following this idea, we propose a novel keypoint concept learning~(KCL) strategy to improve alignment with sparse signals. We introduce a series of new keypoint tokens $\{\langle \boldsymbol{k_i} \rangle\}_{i=1}^N$ to the text encoder according to descriptions $\mathcal{K}=\{\boldsymbol{k}_i\}_{i=1}^N$. These tokens are placeholders of new text embeddings $\mathcal{V}_{kpt} = \{\boldsymbol{v}^*_i\in R^{768}\}_{i=1}^N$, representing new keypoint concepts. These new tokens are appended to captions if the corresponding keypoint concept exists in the pose, {\it i.e.}, $v_i \geq 1$, and we obtain the modified textual prompt $\boldsymbol{c}^\prime_{t}$. Then, we extract spatial attention maps from the cross-attention layers. Given the noisy image $x_t$ and textual prompt $\boldsymbol{c}^\prime_{t}$, the spatial features of the noisy image are projected to a query matrix $Q$, and the text embedding is projected to a key matrix $K$ and a value matrix $V$ via linear projections. The attention map $\mathcal{M}$ is computed as, 
        \begin{equation}
            \label{eq:3-1}
            \mathcal{M}=\text{Softmax}(\frac{Q\cdot K^T}{\sqrt{d}}), 
        \end{equation}
        where $d$ is the latent projection dimension of keys and queries. We select the attention maps of keypoint tokens $\mathcal{M}_{kpt}=\{\mathcal{M}_i\in \mathbb{R}^{H^\prime \times W^\prime}|v_i\geq 1\}$ and encourage them to align with the keypoint heatmap $\mathcal{H}=\{\mathcal{H}_i\in \mathbb{R}^{H^\prime \times W^\prime}| v_i\geq 1\}$~\cite{sun2019hrnet}. This heatmap constraint $\mathcal{L}_{ht}$ is formulated as, 
        \begin{equation}
            \label{eq:3-2}
            \mathcal{L}_{ht}=\frac{1}{|\mathcal{M}_{kpt}|}\cdot\frac{1}{ H^\prime W^\prime}\sum_{v_i\geq 1}||(\mathcal{M}_i - \mathcal{H}_i)||^2, 
        \end{equation}
        where $|\mathcal{M}_{kpt}|$ denotes the number of valid keypoints. As discussed in \S\ref{sec:ablation}, we notice that the attention maps from the $3^{rd}$ transformer block among the $250$$\sim$$500$ time steps are critical for forming these concepts. Empirical studies in \S\ref{sec:discussion} prove that the cross-attention maps of keypoint tokens exhibit high responses spatially aligning with the sparse pose signals, thereby improving the pose alignment with sparse pose signals.
                                             
    \subsection{Overall Pipeline}
    \label{sec:overall}
        As illustrated in Fig.~\ref{fig:arch}, our {\it SP-Ctrl} is built on SD~\cite{Rombach_2022_CVPR_ldm}, consisting of three main components: 1) a {\it spatial embedding module} to render the spatial pose representation $\boldsymbol{c}_{p}$, 2) a {\it text encoder} to encode textual prompts $\boldsymbol{c}^\prime_{t}$ with learnable text embeddings of keypoints, 3) a {\it ControlNet adapter}~\cite{zhang2023adding} to inject spatial conditions into frozen SD model.

        Specifically, the spatial pose representation $\boldsymbol{c}_p$ is first encoded through a conditional embedding module into the latent space. The ControlNet adapter $\mathcal{F}(\cdot;\Theta)$, a trainable copy of the SD encoder, modulates the spatial signals $\boldsymbol{c}_p$ into the diffusion model. Supposing one encoder block $F(\cdot;\theta)$ of SD parameterized by $\theta$, it accepts input feature $\boldsymbol{x}$ and output $\boldsymbol{y}=F(\boldsymbol{x};\theta)$. The ControlNet block $F(\cdot;\theta^\prime)$ with parameter $\theta^\prime$ injects spatial condition $\boldsymbol{c}_p$ with zero convolution to the SD decoder blocks, which is formulated as,  
        \begin{equation}
            \boldsymbol{y_c} = F(\boldsymbol{x};\theta) + \lambda * Z(F(\boldsymbol{x}+Z(\boldsymbol{c}_p;w_2);\theta^\prime);w_1), 
            \label{eq:add-condition}
        \end{equation}
        where $Z(\cdot;w_1)$ and $Z(\cdot;w_2)$ are two $1\times 1$ zero convolution layers with parameters $w_1, w_2$ initialized with zeros for stable training. $\boldsymbol{y_c}$ is the output feature modulated by spatial pose signals. $\lambda$ is the {\it conditioning scale} to adjust the condition strength for inference. 
        
        We run the diffusion model in $\epsilon$-prediction mode, which predicts the noise $\epsilon$ from the noisy latent $\boldsymbol{x}_t$ at time step $t$. While training, we optimize the spatial pose embedding module $\mathcal{G}(\cdot;\phi)$, learnable keypoint embeddings $\mathcal{V}_{kpt}$, and the ControlNet adapter $\mathcal{F}(\cdot;\Theta)$ jointly to minimize a combination of the denoising loss $\mathcal{L}_{ldm}$ and heatmap loss $\mathcal{L}_{ht}$. This process is formulated as,  
        \begin{equation}
            \label{eq:4-1}
            \phi^*, \boldsymbol{V}_{kpt}^*, \Theta^* = \text{arg} \min_{\phi, v_i^*, \Theta} \ \ \mathcal{L}_{ldm} + \eta \cdot\mathcal{L}_{ht}, 
        \end{equation}
        where $\eta$ is a coefficient to balance the two loss items. The denoising loss $\mathcal{L}_{ldm}$ is defined as, 
        \begin{equation}
            \label{eq:4-2}
            \mathcal{L}_{ldm} = \mathbb{E}_{\boldsymbol{x},y,\epsilon,t}[||\epsilon - \epsilon(\boldsymbol{x}_t, t, \boldsymbol{c}_{p}, \boldsymbol{c}^\prime_{t}; \Theta)||^2], 
        \end{equation}
        where $\epsilon(\cdot)$ is the denoising network. Since we observe appearance collapse when introducing learnable text embeddings $\mathcal{V}_{kpt}$, we detach the gradient from the noisy image query $Q$ for $\mathcal{L}_{ht}$ to avoid potential information leakage. 

\section{Experiments}
\label{sec:experiments}
    \subsection{Datasets and Metrics}

        \para{AP-10K.} AP-10K~\cite{yu2021ap} dataset contains $10,105$ images collected and filtered from 23 animal families and 54 species of mammals. It defines 17 keypoints sharing across all species (see Appendix \S B). We adopt the training set for training and conduct evaluation on the validation set. Since AP-10K does not contain textual descriptions, we employ several prompt templates including species and background definitions following \cite{Ju_humansd}, as presented in Appendix \S B.  

        \para{Human-Art.} Human-Art~\cite{ju2023humanart} dataset comprises $50,000$ high-quality human-centric images from 5 real-world and 15 virtual scenarios, including both 2D and 3D scenarios. It provides bounding boxes, keypoint annotations, and textual descriptions. We employ the 17 keypoint definitions in experiments, as presented in Appendix \S B. We use the training set for training and report the metrics on the validation set.     

        \para{Metrics.} The evaluation metrics include: 1) {\it Pose mAP.} We employ the SOTA pose estimator ViTPose++-H~\cite{xu2022vitpose+}, pretrained on AP-10K and Human-Art~\cite{mmpose2020}, to detect the keypoints of generated images and report OKS-based mAP metric~\cite{ruggero2017benchmarking} to evaluate pose alignment. The higher mAP means better pose alignment. 2) {\it CLIP-Score.} We report the category-averaged CLIP-Score~(ViT-L/14)~\cite{hessel2021clipscore} to measure the image-text alignment. 3) {\it FID.} We employ FID to evaluate the image fidelity. 4) {\it Detection AP$.75$.} We utilize an open-set object detector, Grounding-DINO~\cite{liu2024grounding}, to detect the objects in the images by prompting the category name, which reflects the appearance fidelity of generated images. Considering the possible variance of the objects' shape and positions, especially when using sparse signals for pose guidance, we report the AP$.75$ for reference. 

        \makeatletter
            \newcommand{\ssymbol}[1]{^{\@fnsymbol{#1}}}
        \makeatother

        \definecolor{GrayBackground}{gray}{0.9}
        \begin{table}[b]
            \centering
            \renewcommand\arraystretch{1.2}
            \scalebox{0.82}{
            \begin{tabular}{c|c|cccc}
                \toprule
                \multirow{2}{*}{Dataset} & \multirow{2}{*}{Method} & Pose & \multirow{2}{*}{FID$\downarrow$} & CLIP- &  Detection \\
                ~ & ~ & mAP$\uparrow$ & ~ & Score$\uparrow$ &  AP.75$\uparrow$ \\
                \midrule
                \multirow{4}{*}{AP-10K} & Real Image & 83.00 & — & 24.45 & 50.81  \\
                \cline{2-6}
                ~ & T2I-Adapter & 48.16 & 27.29 & {\bf 25.52} & 24.23 \\
                ~ & ControlNet & 44.25 & 19.40 & 24.77 & 24.35 \\
                ~ & \cellcolor{GrayBackground} {\bf Ours} & \cellcolor{GrayBackground} {\bf 55.63} & \cellcolor{GrayBackground} {\bf 18.52} & \cellcolor{GrayBackground} 23.86 & \cellcolor{GrayBackground} {\bf 25.10} \\
                \bottomrule

                \midrule
                \multirow{6}{*}{Human-Art} & Real Image & 79.97 & - & 26.87 & 6.78   \\
                \cline{2-6}
                ~ & T2I-Adapter & 38.37 & 40.01 & 27.36 & 6.78 \\
                ~ & ControlNet & 45.26 & {\bf 26.69} & 27.84 & 8.18 \\
                ~ & HumanSD$\ssymbol{2}$ & 49.92 & 35.18 & 27.35 & 8.29 \\
                ~ & GRPose$\ssymbol{2}$ & 50.93 & 28.85 & {\bf 27.95} & 6.51 \\
                ~ & \cellcolor{GrayBackground} {\bf Ours} & \cellcolor{GrayBackground} {\bf 51.11} & \cellcolor{GrayBackground} 29.30 & \cellcolor{GrayBackground} 25.94 & \cellcolor{GrayBackground} {\bf 9.11}  \\
                \bottomrule
            \end{tabular}}
            \caption{Performance comparison with other popular pose-guided T2I methods on AP-10K and Human-Art dataset. $\ssymbol{2}$ means additional pretrained pose estimators are required when training.}
            \label{tab:baseline-compare}
        \end{table}

    \subsection{Implementation Details}
        We adopt a frozen SD v1.5~\cite{sd_v15} as the base model, and the ControlNet adapter is trained from scratch using SD encoder parameters for initialization in all experiments. The learning rate is set to 1$e$-5. We randomly drop $50\%$ prompts for classifier-free guidance. We initialize the keypoint embeddings $\boldsymbol{E}_0$ with randomly fixed vectors and select the cross-attention maps from the $3^{rd}$ transformer block among the $250$$\sim$$500$ time steps for computing $\mathcal{L}_{ht}$ as discussed in \S\ref{sec:discussion}. We set $\eta=0.1$ according to the discussions in Appendix \S C. We train the model for 800 epochs for AP-10K and 300 epochs for Human-Art. To inference, we employ a UniPC~\cite{zhao2023unipc} sampler with 50 sampling steps with a CFG scale of $7.5$ and conditioning scale $\lambda=1.0$. No negative prompt is used. We generate three images for each prompt with different sampled initial noise for evaluation. All experiments are conducted on \text{NVIDIA A100 40G GPUs}.

    \subsection{Comparison with Other Methods}
        \begin{figure*}[t]
            \centering
            \includegraphics[width=0.94\linewidth]{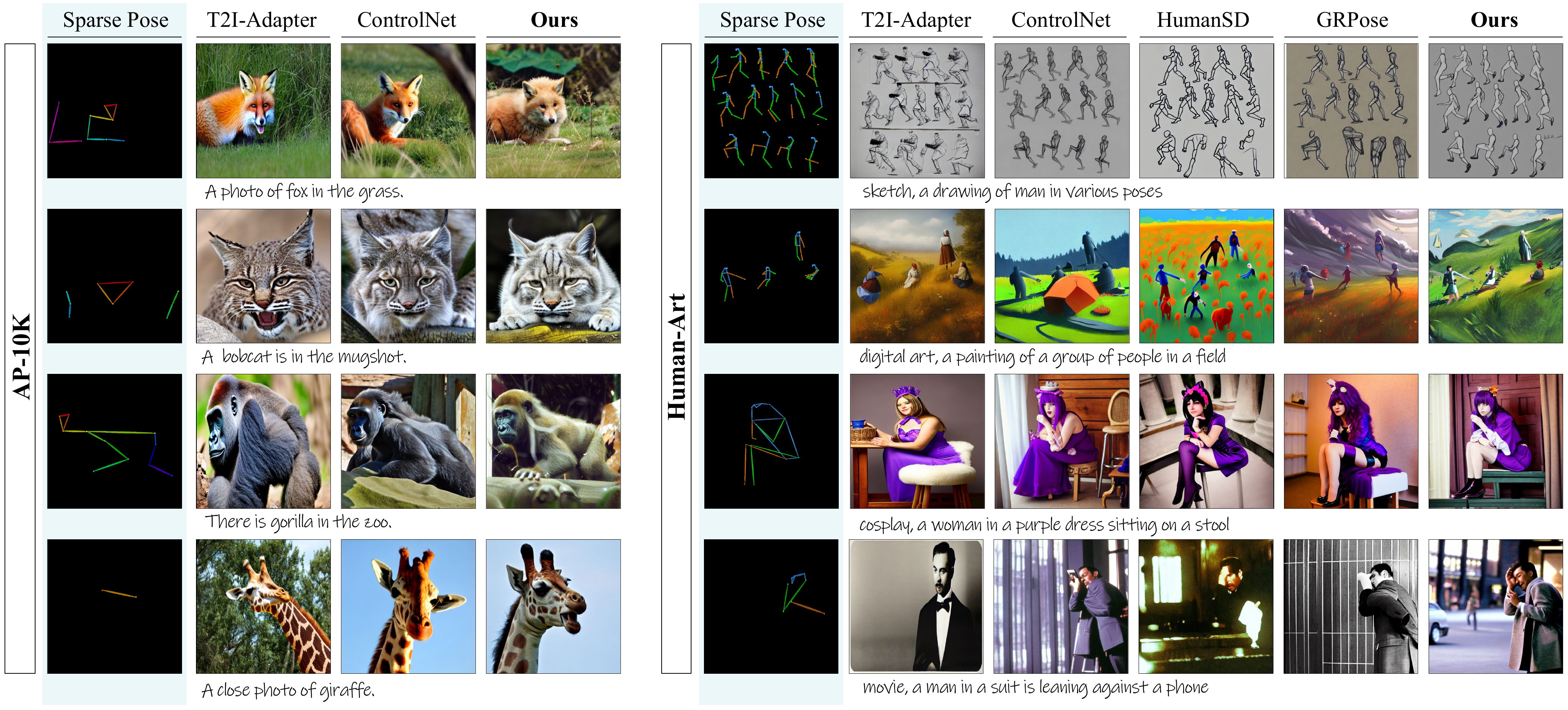}
            \vspace{-0.5em}
            \caption{Visualized comparison of our {\it SP-Ctrl} and other popular methods. See Appendix Fig.~S8 and Fig.~S9 for more examples.}
            \vspace{-1em}
            \label{fig:comparison-with-sota}
        \end{figure*}
    
        \para{Performance on AP-10K.} The performance is reported in Tab.~\ref{tab:baseline-compare}. Our method obtains the best pose mAP and FID, outperforming the baseline ControlNet by $11.38\%$ and $0.88$. Since the pretrained CLIP model's vocabulary does not contain the newly introduced keypoint tokens $\mathcal{V}_{kpt}$ introduced in \S\ref{sec:kcl}, we remove such tokens to compute the CLIP-Score for comparison. This operation slightly reduces the CLIP-Score due to inference-evaluation discrepancies, but we find it acceptable as discussed in \S\ref{sec:ablation} for Fig.~\ref{fig:discussion-kcl}. 
        The competitive detection AP$.75$ indicates the faithfulness to animal species. Visualized examples are presented in Fig.~\ref{fig:comparison-with-sota}. 
        Our method can discriminate the keypoint and align with detailed pose strictly, such as paws and nose, while other methods may misinterpret some keypoints and cause misalignment. 
    
        \para{Performance on Human-Art.} In Tab.~\ref{tab:baseline-compare}, our method also achieves promising pose control over human-centric generation tasks with the best pose mAP $51.11\%$—matching the performance of the SOTA GRPose—even without introducing extra pretrained pose estimators for computing perceptive losses during training. Moreover, our method brings little additional computational cost to inference compared with ControlNet. The decrement of CLIP-Score is acceptable as discussed in \S\ref{sec:ablation}. The marginal discrepancy in FID is supposed to be negligible, and the resulting quality remains high in Fig.~\ref{fig:comparison-with-sota}. Note that our method produces images with distinguishable and well-aligned poses, excelling at local structures like arms and legs. This proves the effectiveness and advantages of our method. 

    \subsection{Ablation Studies}
    \label{sec:ablation}
        All experiments in this section were conducted on the AP-10K dataset. The baseline is ControlNet with OpenPose. 

        \begin{figure}[t]
            \centering
            \includegraphics[width=0.90\linewidth]{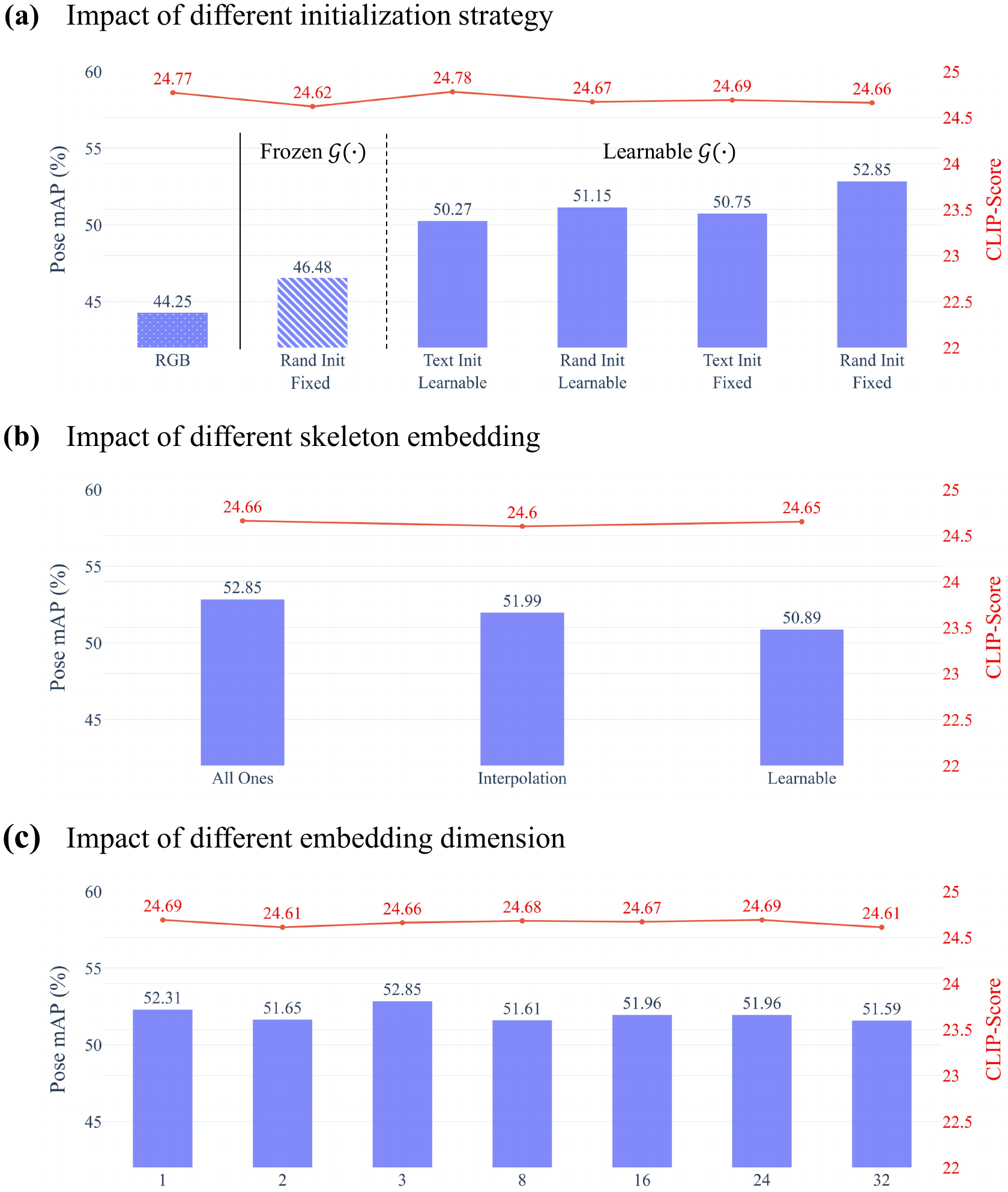}
            \vspace{-0.5em}
            \caption{Discussions on the spatial-pose representation.
            }
            \vspace{-1.0em}
            \label{fig:discussion-spe}
        \end{figure}
        
        \para{Discussions on spatial-pose representation.} We report the results in Fig.~\ref{fig:discussion-spe}, including three parts as follows. 

        {\it Effects of initialization of $\boldsymbol{E}_0$.} An intuitive idea is using text embeddings of keypoint descriptions to initialize $\boldsymbol{E}_0$($C$=$768$), which introduces semantic priors explicitly~\cite{avrahami2023spatext}. We compare this setting with random initialization and compare the learnable and fixed $\boldsymbol{E}_0$ when training. By default, we set the skeleton embedding $\boldsymbol{e}_0=\boldsymbol{1}^{1\times C^\prime}$ and the keypoint embedding dimension $C^\prime$$=$$3$. The baseline is the RGB embeddings of OpenPose. In Fig.~\ref{fig:discussion-spe}~(a), the RGB embeddings show the lowest pose mAP, even lower than randomly initialized $\boldsymbol{E}_0$ with a frozen embedding module $\mathcal{G}(\cdot)$, indicating that OpenPose even confuses pose guidance. Using a learnable $\mathcal{G}(\cdot)$ to optimize the keypoint embeddings $\boldsymbol{E}_{kpt}$ brings remarkable improvement on mAP. When we set $\boldsymbol{E}_0$ as learnable, using randomly initialized $\boldsymbol{E}_0$ shows $0.88\%$ higher mAP than that initialized with text embeddings, and so does the fixed $\boldsymbol{E}_0$. This implies possible discrepancies between text-embedding spaces and spatial-pose representations, and the random initialization provides approximately orthogonal vectors for optimizing keypoint embeddings. Meanwhile, fixed $\boldsymbol{E}_0$ brings consistently better mAP than the learnable one owing to optimization stability. The difference in CLIP-Score is negligible. Thus, we employ fixed, randomly initialized $\boldsymbol{E}_0$ with learnable $\mathcal{G}(\cdot)$. 

        {\it Effects of different skeleton embedding $\boldsymbol{e}_{sks}$.} We discuss the impact of different skeleton embedding $\boldsymbol{e}_{sks}$, including {\bf 1)} all-one embedding, {\bf 2)} interpolation of adjacent keypoint embeddings, and {\bf 3)} learnable embeddings. As shown in Fig.~\ref{fig:discussion-spe}~(b), it achieves the best mAP when simply setting $\boldsymbol{e}_{sks}$ to all ones, which is $0.86\%$ higher than interpolation and $1.96\%$ higher than learnable vectors. We suppose that the all-ones vector is enough to provide useful spatial guidance like the topology of poses. The difference in CLIP-Score is negligible. So, we set $\boldsymbol{e}_{sks}=\mathbf{1}^{1\times C^\prime}$ by default. 

        {\it Effects of embedding dimension $C^\prime$.} We further explore the embedding dimension $C^\prime$ of spatial-pose representation. As reported in Fig.~\ref{fig:discussion-spe}~(c), we surprisingly find that different settings of $C^\prime$ exhibit competitive performance on both pose mAP and CLIP-Sore, and even one-channel keypoint embedding can obtain competitive performance. This reflects the expressiveness of our spatial-pose representation. 
        To compare with Openpose, we adopt $C^\prime=3$ for consistency. Note that the higher embedding dimension $C^\prime$ would cause increased computation in the condition embedding module.  

        \begin{figure}[t]
            \centering
            \includegraphics[width=0.90\linewidth]{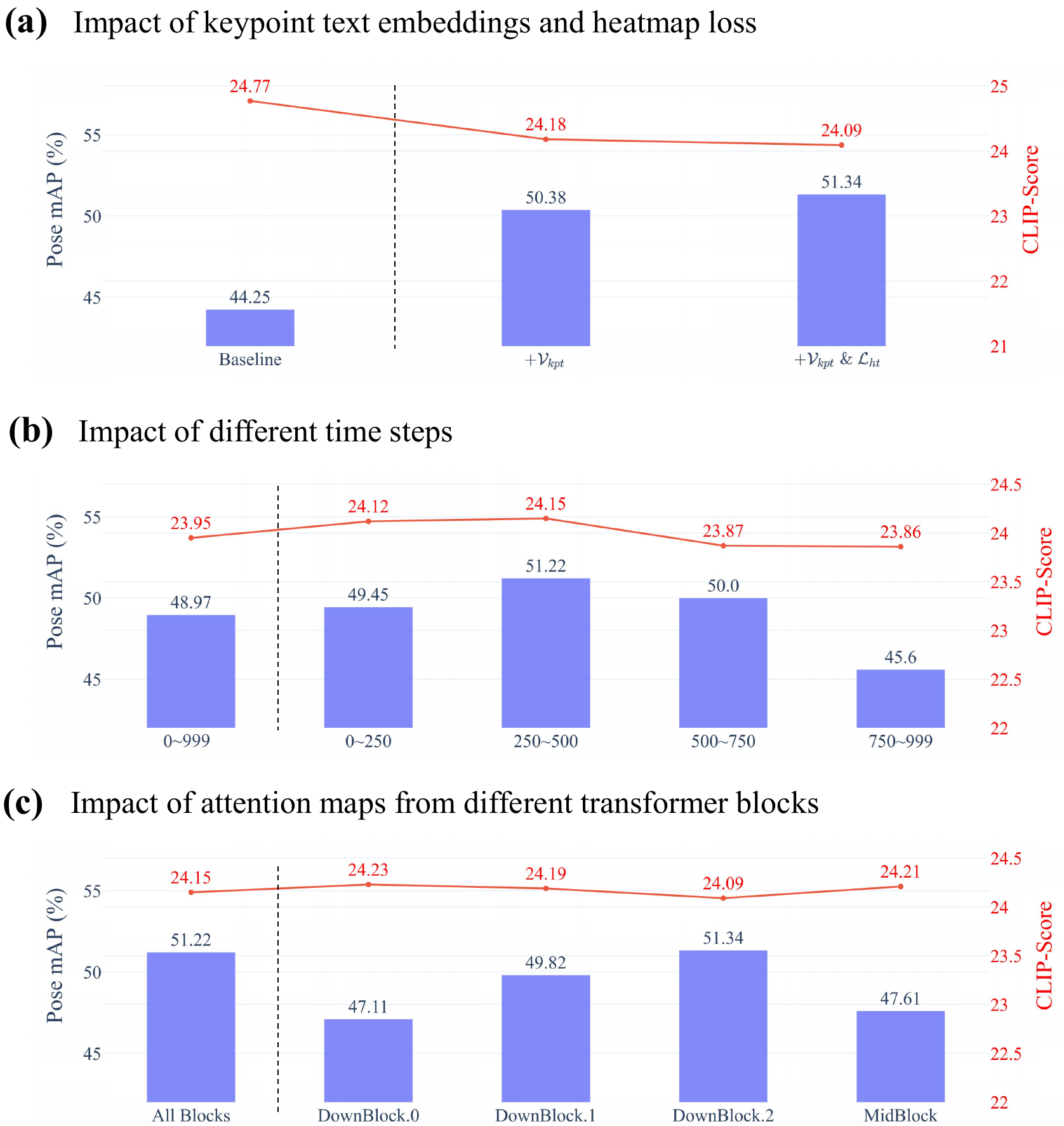}
            \vspace{-0.5em}
            \caption{Discussions on the keypoint concept learning strategy.
            }
            \vspace{-1.5em}
            \label{fig:discussion-kcl}
        \end{figure}
        
        \para{Discussions on keypoint concept learning.} We explore the functions of new keypoint tokens $\mathcal{V}_{kpt}$ and heatmap loss $\mathcal{L}_{ht}$ with OpenPose, which is presented in Fig.~\ref{fig:discussion-kcl}. 

        {\it Effects of keypoint embeddings $\mathcal{V}_{kpt}$ and heatmap loss $\mathcal{L}_{ht}$.} As shown in Fig.~\ref{fig:discussion-kcl}~(a), the $\mathcal{V}_{kpt}$ alone improves mAP by $6.13\%$, which learns the keypoint concepts implicitly through the existence of keypoints. With the heatmap constraint $\mathcal{L}_{ht}$, our method obtains further increments, achieving the best mAP at $51.34\%$. Given that the pretrained CLIP model's vocabulary does not contain the new tokens of $\mathcal{V}_{kpt}$, we remove such tokens to compute CLIP-Score for comparison, which leads to a slightly reduced CLIP-Score ($0.68$$\downarrow$) due to inference-evaluation discrepancies. We further check the metrics of those generated without keypoint tokens, where CLIP-Score is $24.94$ and mAP is $45.07\%$. See Appendix \S C for samples. This validates the importance of keypoint embeddings $\mathcal{V}_{kpt}$. Since the visualized results in Fig.~\ref{fig:comparison-with-sota} maintain faithful text-image alignments, we suppose the subtle decrease of CLIP-Score is acceptable.

        {\it Effects of different time steps.} As shown in Fig.~\ref{fig:discussion-kcl}~(b), different time steps contribute differently to learning new keypoint concepts with $\mathcal{L}_{ht}$ during the diffusion process. The $250$$\sim$$500$ time steps are the most important, and the $750$$\sim$$999$ time steps are useless. The model also achieves the best CLIP-Score when implementing $\mathcal{L}_{ht}$ at $250$$\sim$$500$ time steps. This observation provides valuable insight into the role of time steps in forming the concepts of subjects during the denoising process~\cite{balaji2022ediff, mou2024t2i}.
        
        {\it Effects of attention maps from different transformer blocks.} We further explore the functions of attention maps from different transformer blocks of the denoising U-Net encoder, especially those from the $250$$\sim$$500$ time steps. As shown in Fig.~\ref{fig:discussion-kcl}~(c), the $3^{rd}$ transformer block of the denoising U-Net encoder, {\it i.e.}, \texttt{DownBlock.2}, contributes mostly, which achieves almost the same pose mAP as that using all blocks. This shows the crucial role of the $3^{rd}$ transformer block in forming the new keypoint concepts. The discrepancy of CLIP-Score is negligible. Thus, we adopt cross-attention maps from \texttt{DownBlock.2} during the $250$$\sim$$500$ time steps for computing $\mathcal{L}_{ht}$.

        \para{Ablations for the proposed modules.} The results are listed in Tab.~\ref{tab:ablation-modules}. Both proposed modules demonstrate effectiveness compared to the baseline, where spatial-pose representation improves pose mAP by $8.60\%$, and keypoint concept learning increases $7.09\%$. Combining both, our {\it SP-Ctrl} achieves the best pose mAP $55.63\%$, showing the compatibility of our two designs. Our method also obtains the best FID. 
    
        \begin{table}[t]
            \centering
            \renewcommand\arraystretch{0.87}
            \scalebox{0.96}{
            \begin{tabular}{l|cccc}
                \toprule
                 Method & Pose mAP$\uparrow$ & FID$\downarrow$ & CLIP-Score$\uparrow$ \\
                \midrule
                \textbf{ControlNet} & 44.25 & 19.40 & {\bf 24.77}  \\
                \midrule
                + Spatial Pose & 52.85 & 19.67 & 24.62 \\ 
                + KCL & 51.34 & 18.94 & 24.09 \\
                \midrule
                \rowcolor{GrayBackground} \textbf{Ours} & {\bf 55.63} & {\bf 18.52} & 23.86 \\ 
                \bottomrule
            \end{tabular}}
            \vspace{-0.5em}
            \caption{Ablations of proposed modules, \ie, the spatial-pose representation and keypoint concept learning strategy.}
            \vspace{-0.5em}
            \label{tab:ablation-modules}
        \end{table}

        \begin{table}[t]
            \centering
            \renewcommand\arraystretch{1.1}
            \scalebox{0.83}{
            \begin{tabular}{c|c|cccc}
                \toprule
                Condition & Method & Pose mAP$\uparrow$ & FID$\downarrow$ & CLIP-Score$\uparrow$ \\
                \midrule
                Mask & ControlNet & 26.74 & 21.15 & 24.39 \\
                \midrule
                Depth & ControlNet & {\bf 57.07} & 19.85 & 24.44 \\ 
                \midrule
                \multirow{2}{*}{Sparse Pose} & ControlNet & 44.25 & 19.40 & {\bf 24.77} \\ 
                ~ & \cellcolor{GrayBackground}{Ours} & \cellcolor{GrayBackground}{55.63} & \cellcolor{GrayBackground}{\bf 18.52} & \cellcolor{GrayBackground}{23.86} \\ 
                \bottomrule
            \end{tabular}}
            \vspace{-0.5em}
            \caption{Performance of different conditional guidance.}
            \label{tab:discuss-conditions}
            \vspace{-1.7em}
        \end{table}

    \subsection{Discussions}
    \label{sec:discussion}

\begin{figure}[t]
            \centering
            \includegraphics[width=0.92\linewidth]{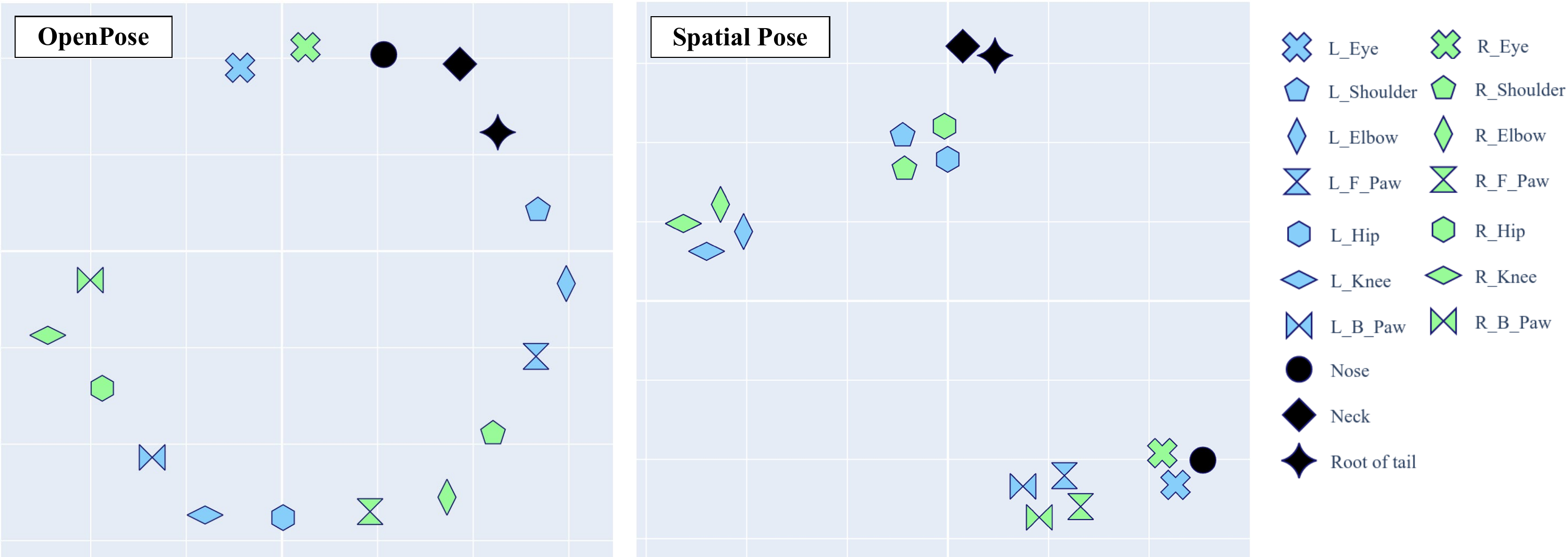}
            \vspace{-0.5em}
            \caption{Comparison of the RGB embeddings and learned keypoint embeddings $\boldsymbol{E}_{kpt}$ after t-SNE~\cite{tsne}.}
            \vspace{-0.5em}
            \label{fig:validation-spe}
        \end{figure}

        \para{Validations of spatial pose.} Fig.~\ref{fig:validation-spe} shows the distribution of RGB embeddings from OpenPose and the learned embeddings $\boldsymbol{E}_{kpt}$ from our spatial-pose representation after t-SNE~\cite{tsne}. Compared to plain RGB embeddings, the learned $\boldsymbol{E}_{kpt}$ pulls semantic-similar and symmetric keypoints together, thus encoding semantic priors into the channels of sparse signals and making each keypoint expressive. This proves the discriminative guidance of our representation. 

\begin{figure}[t]
            \centering
            \includegraphics[width=0.92\linewidth]{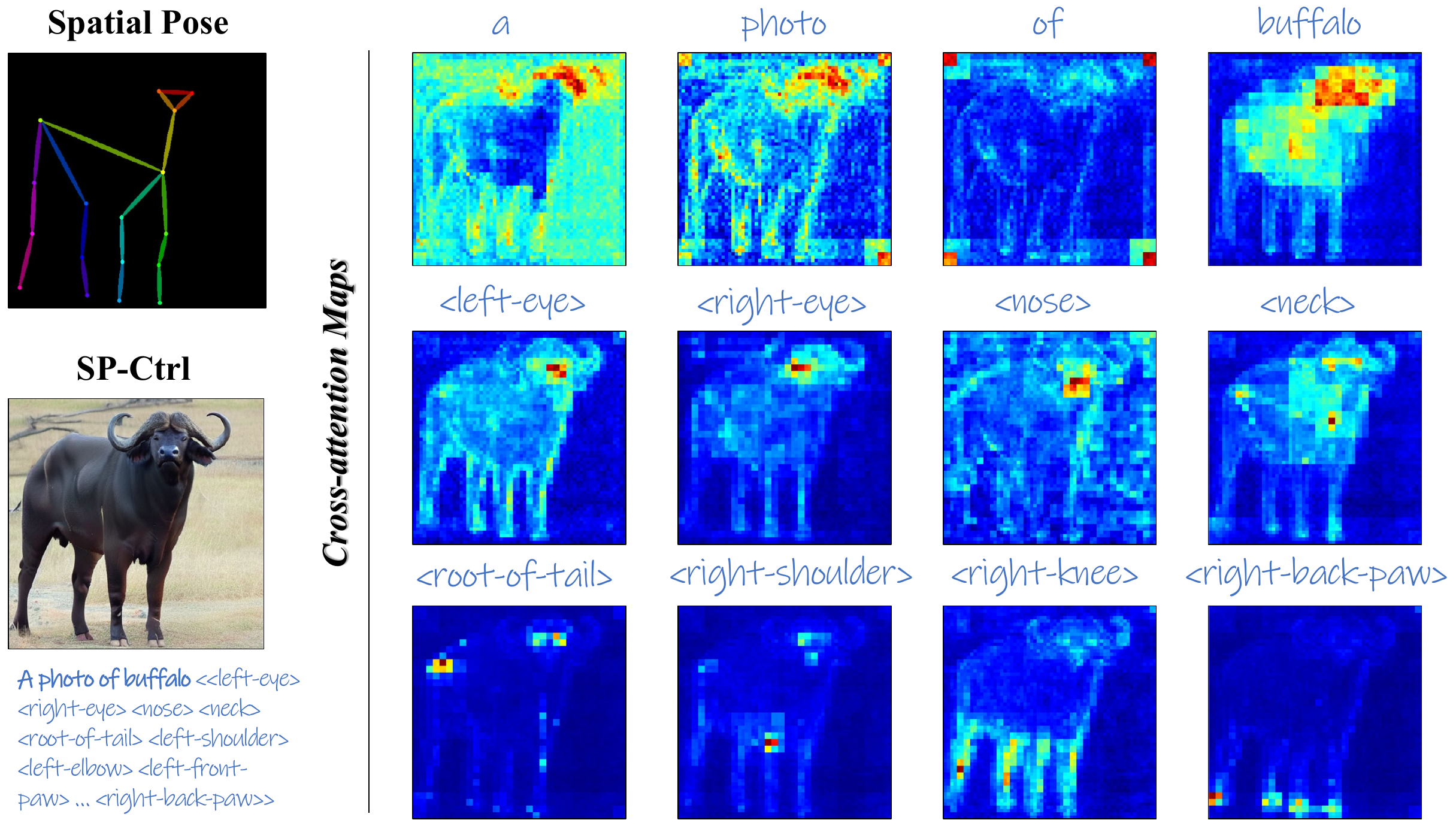}
            \vspace{-0.5em}
            \caption{Visualized cross-attention maps of {\it SP-Ctrl} at time step $399$, where the net keypoint tokens $\{\langle \boldsymbol{k_i} \rangle\}_{i=1}^N$ attend to spatial positions of keypoints. Full results refer to Appendix Fig.~S7.}
            \vspace{-0.5em}
            \label{fig:validation-kcl}
        \end{figure}

        \para{Validations of keypoint tokens.} We check the cross-attention maps of our method in Fig.~\ref{fig:validation-kcl}, especially for these introduced keypoint tokens $\{\langle \boldsymbol{k_i} \rangle\}_{i=1}^N$. Under the keypoint-heatmap guidance, these new tokens show high spatial responses at the keypoint positions, validating the effectiveness of keypoint concept learning. It encourages pose alignment and increases focus on spatial pose signals.   

\begin{figure}[t]
            \centering
            \includegraphics[width=0.92\linewidth]{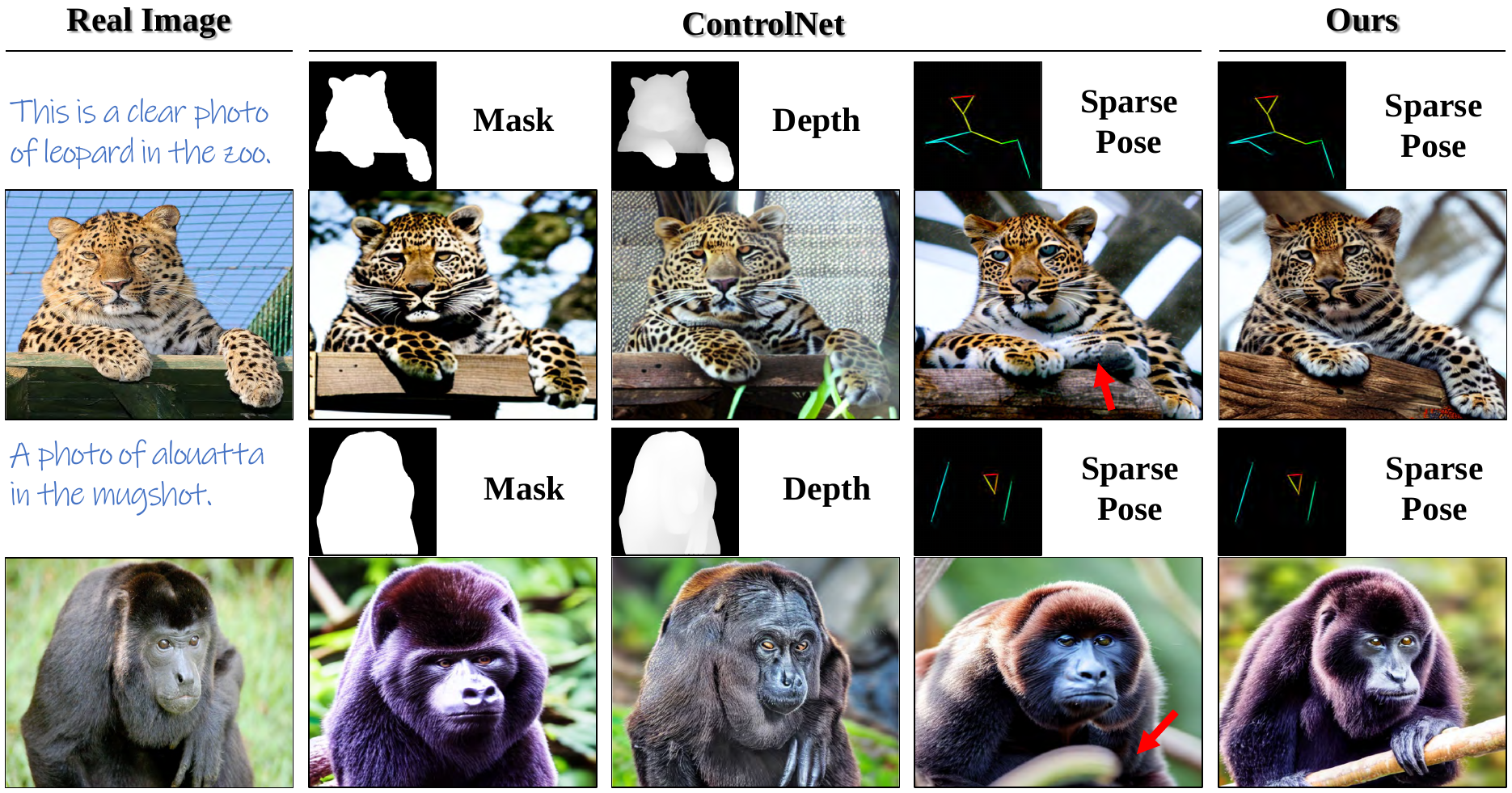}
            \vspace{-0.5em}
            \caption{Examples of different guidance, where ours exhibits high pose alignment. More examples refer to Appendix Fig.~S10.}
            \vspace{-1.3em}
            \label{fig:different-condition-vis}
        \end{figure}

        \para{Comparison with other conditions.} Tab.~\ref{tab:discuss-conditions} reports the results of pose-controllable generation through various signals including masks, depth, and OpenPose. We employ SAM-HQ~\cite{sam_hq} to extract semantic masks and ZoeDepth~\cite{bhat2023zoedepth} for depth maps. Our method surpasses the baseline ControlNet by $11.38\%$ mAP, almost matching the performance of using depth maps. The semantic masks fails to control the exact pose. Visualized results are shown in Fig.~\ref{fig:different-condition-vis}.

\begin{figure}[t]
            \centering
            \includegraphics[width=0.96\linewidth]{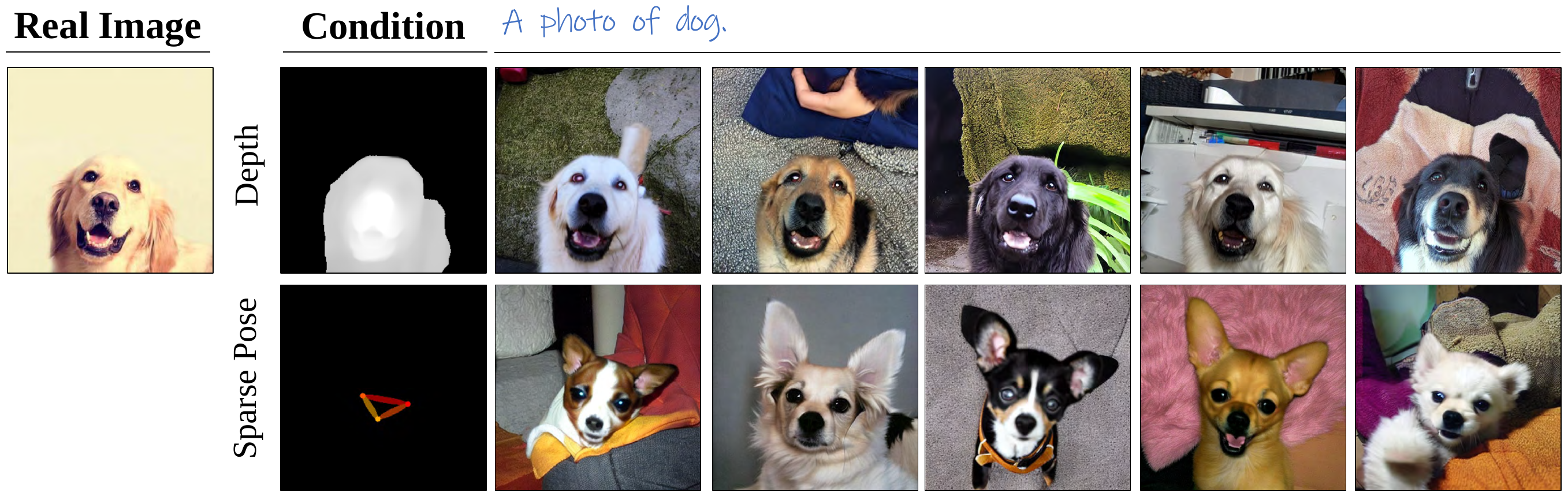}
            \vspace{-0.5em}
            \caption{Examples of shape diversity, where our method synthesizes images with precisely aligned poses and diverse shapes. More examples refer to Appendix Fig.~S11.}
            \vspace{-0.5em}
            \label{fig:show-diversity}
        \end{figure}

\begin{figure}[t]
            \centering
            \includegraphics[width=0.96\linewidth]{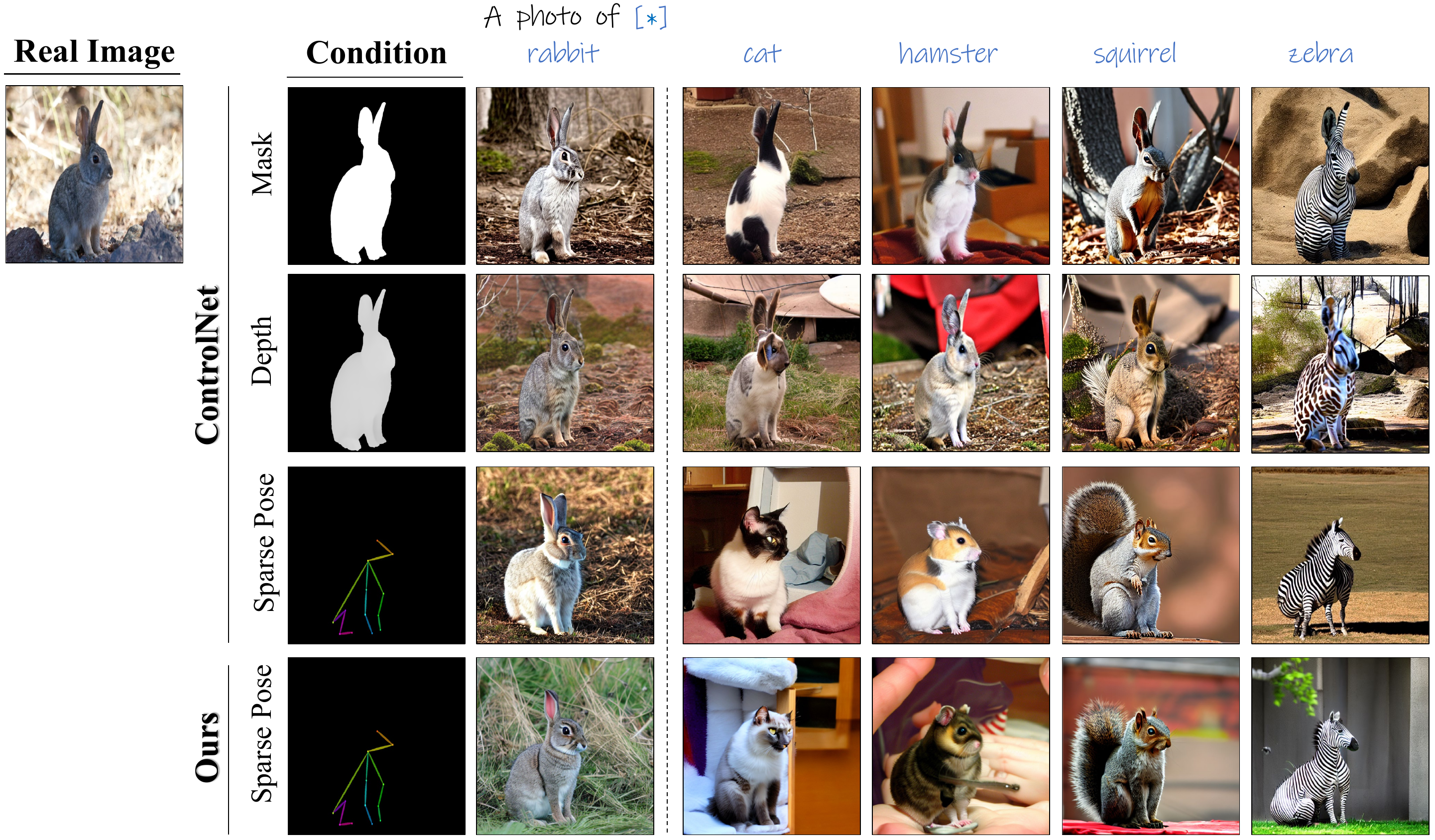}
            \vspace{-0.5em}
            \caption{Examples of cross-species pose-guided generation results. More examples refer to Appendix Fig.~S12.}
            \vspace{-0.5em}
            \label{fig:show-cross-species-gen}
        \end{figure}

        \para{Advantages of sparse-pose guidance.} We showcase the potential of sparse-pose signals in several application scenarios. Given the precise pose control with sparse pose, our SP-Ctrl enjoys diverse synthesized images in various shapes~(Fig.~\ref{fig:show-diversity}) and cross-species generation capability~(Fig.~\ref{fig:show-cross-species-gen}). Due to the simplicity of sparse pose, our method also offers high flexibility in pose editing and allows the creation~(Fig.~\ref{fig:show-editing}) of new poses at a low cost. Additional examples are presented in Appendix \S D.

        \begin{figure}[t]
            \centering
            \includegraphics[width=0.98\linewidth]{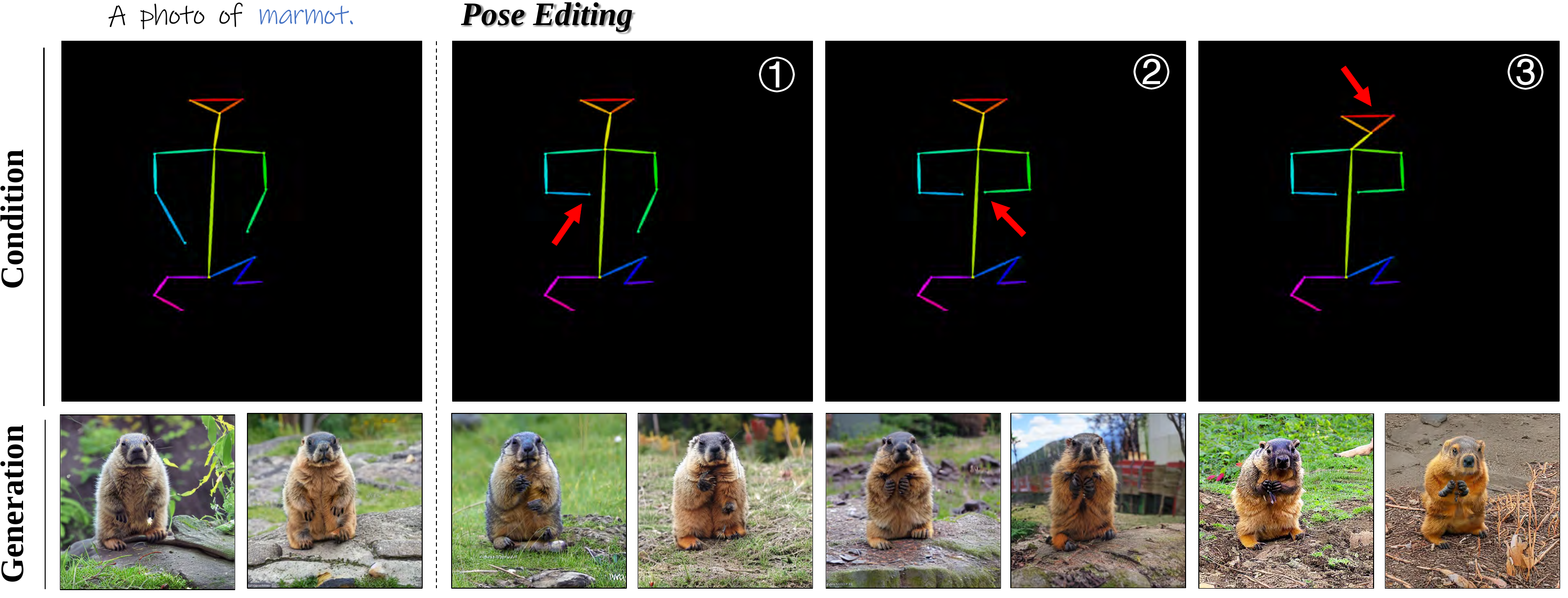}
            \vspace{-0.5em}
            \caption{Examples of synthesized images with edited sparse pose signals. More examples refer to Appendix Fig.~S13.}
            \vspace{-1.0em}
            \label{fig:show-editing}
        \end{figure}

\section{Conclusion}
\label{sec:conclusion}
    This paper rethinks sparse signals, \ie, OpenPose, for pose-guided text-to-image generation and proposes {\it SP-Ctrl}, which realizes precise pose control with sparse signals. To cover the limited guidance of sparse signals, a learnable spatial pose representation is first developed to encode keypoint priors into pose guidance. Additionally, we design a novel keypoint concept learning strategy, which achieves improved pose alignment by enhancing the spatial response of each keypoint token on the cross-attention maps. Extensive experiments demonstrate the effectiveness of our method in pose-controllable generation, even matching the performance of dense signal-based methods. Moreover, our method exhibits the advantages of sparse pose signals in image diversity, cross-species generalization, and pose editing.  

{
    \small
    \bibliographystyle{ieeenat_fullname}
    \bibliography{main}
}

\appendix
\renewcommand{\thefigure}{S\arabic{figure}}
\renewcommand{\thetable}{S\arabic{table}}

\section*{Appendix}

    This appendix is organized as follows.
    
    \begin{itemize}
        \item Implementation details on the model structure. (\S\ref{sec:app-spe})
        \item More details about datasets and keypoint definition. (\S\ref{sec:app-keypoint-definition})
        \item More discussions are provided, including the effects of loss coefficient $\eta$ and an analysis of the decrement on the CLIP-Score. (\S\ref{sec:app-discussion})
        \item Additional visualized examples with detailed illustrations are presented to supplement the main paper. (\S\ref{sec:app-more-results})
        \item A discussion about limitations and future works. (\S\ref{sec:app-limitations})
    \end{itemize}

    \begin{table}[h]
        \centering
        \scalebox{0.80}{
        \begin{tabular}{ccl}
            \toprule
            \bf{ID}  & \makecell{\bf{Reference} \\(main paper)} & {\bf Brief Illustration} \\
            \midrule
            Fig.~\ref{fig:app-spe} & \S{\bf3.1}, Line 191  & \makecell[l]{Detailed architecture of the spatial\\  embedding module $\mathcal{G}(\cdot)$.} \\
            \midrule
            Fig.~\ref{fig:app-kpt-definition} & \makecell[l]{\S{\bf3.1}, Line 170,\\ \S{\bf4.1}, Line 291,301} &  \makecell[l]{Keypoint descriptions of AP-10K\\ and Human-Art dataset.} \\
            \midrule
            Fig.~\ref{fig:app-ablation-loss-scale} & \S{\bf4.2}, Line 328 &  \makecell[l]{Discussions on loss coefficient $\eta$.} \\
            \midrule
            Fig.~\ref{fig:app-clipscore-analysis} & \S{\bf4.4}, Line 427 &  \makecell[l]{Analysis of the CLIP-Score.} \\
            \midrule
            Fig.~\ref{fig:app-limitation} & -- &  \makecell[l]{Illustrations on the imitations.} \\
            \midrule
            Fig.~\ref{fig:app-controlnet-cross-attention-maps} & \S{\bf3.2}, Fig.~4 &  \makecell[l]{Full cross-attention maps of\\ ControlNet.} \\
            \midrule
            Fig.~\ref{fig:app-spctrl-cross-attention-maps} & \S{\bf4.5}, Fig.~9 &  \makecell[l]{Full cross-attention maps of our\\ SP-Ctrl.} \\
            \midrule
            Fig.~\ref{fig:app-comparison-ap10k} & \S{\bf4.3}, Fig.~5 & \makecell[l]{More examples on AP-10K.} \\
            \midrule
            Fig.~\ref{fig:app-comparison-humanart} & \S{\bf4.3}, Fig.~5 & \makecell[l]{More examples on Human-Art.} \\
            \midrule
            Fig.~\ref{fig:app-different-condition} & \S{\bf4.5}, Fig.~10 & \makecell[l]{More examples of generation\\ with different conditions.} \\
            \midrule
            Fig.~\ref{fig:app-diverse-shape} & \S{\bf4.5}, Fig.~11 & \makecell[l]{More examples to show the\\ shape diversity of our method.} \\
            \midrule
            Fig.~\ref{fig:app-cross-species} & \S{\bf4.5}, Fig.~12 & \makecell[l]{More examples to show the cross-\\species generation.} \\
            \midrule
            Fig.~\ref{fig:app-editing} & \S{\bf5.5}, Fig.~13 & \makecell[l]{More examples to show the pose-\\editing results.} \\
            \midrule
            Tab.\ref{tab:app-prompt-template} & \S{\bf4.1}, Line 295 & \makecell[l]{Prompt templates for AP-10K.} \\
            \bottomrule
        \end{tabular}}
        \caption{Quick overview of figures and tables in the Appendix.}
        \label{tab:fig-index}
    \end{table}

    \section{More Details of Model Structure}
    \label{sec:app-spe}
\begin{figure}[t]
            \centering
            \includegraphics[width=0.95\linewidth]{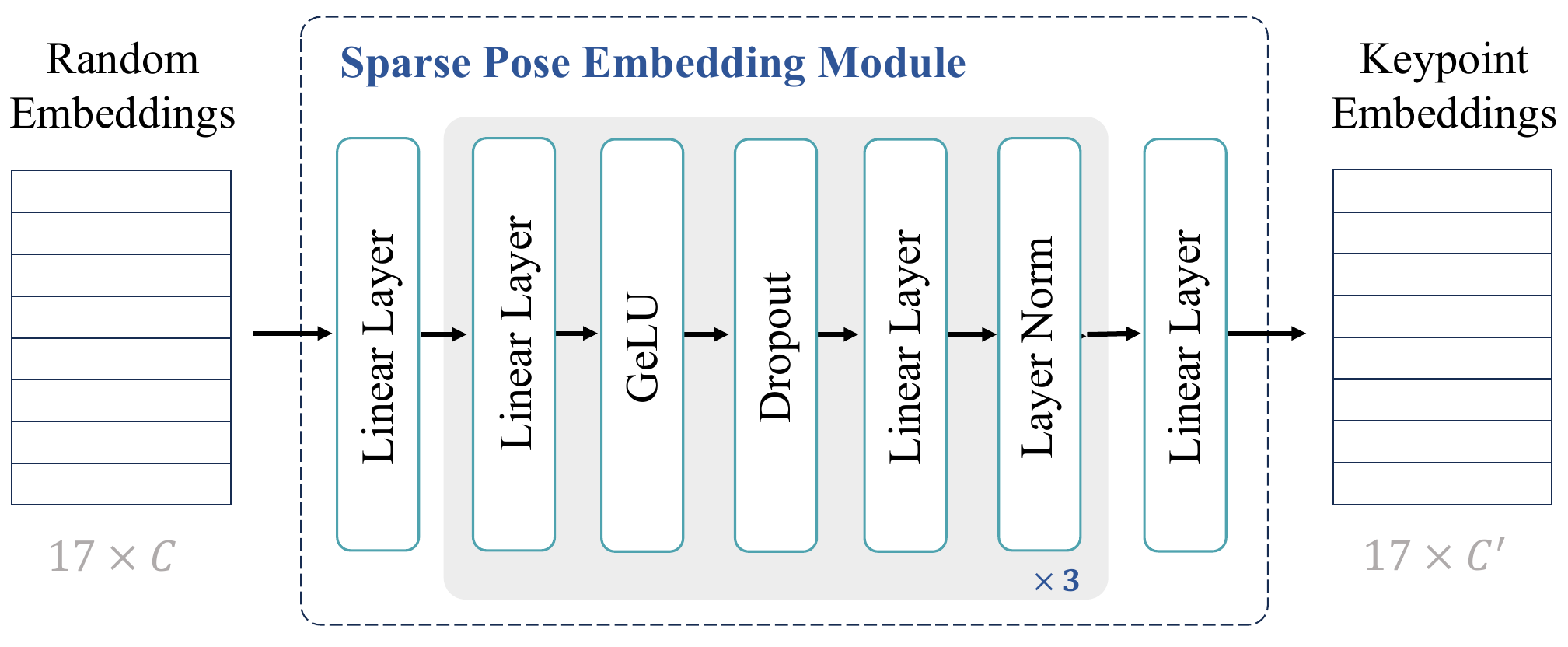}
            \caption{The detailed structure of our sparse pose embedding module $\mathcal{G}(\cdot)$. }
            \label{fig:app-spe}
        \end{figure}

        We provide the detailed structure of our sparse-pose embedding module $\mathcal{G}(\cdot)$ in Fig.~\ref{fig:app-spe}. It comprises two linear layers and three basic blocks of stacked \texttt{Linear + GeLU + Dropout + Linear + Layer Norm} layers. The module accepts the random initialized vector $\boldsymbol{E}_0$ and outputs the learned keypoint embeddings $\boldsymbol{E}_{kpt}$ for constructing the spatial-pose representation, which is optimized directly for the denoising diffusion objective.

    \section{More Details of Datasets}
    \label{sec:app-keypoint-definition}

\begin{figure}[t]
            \centering
            \includegraphics[width=\linewidth]{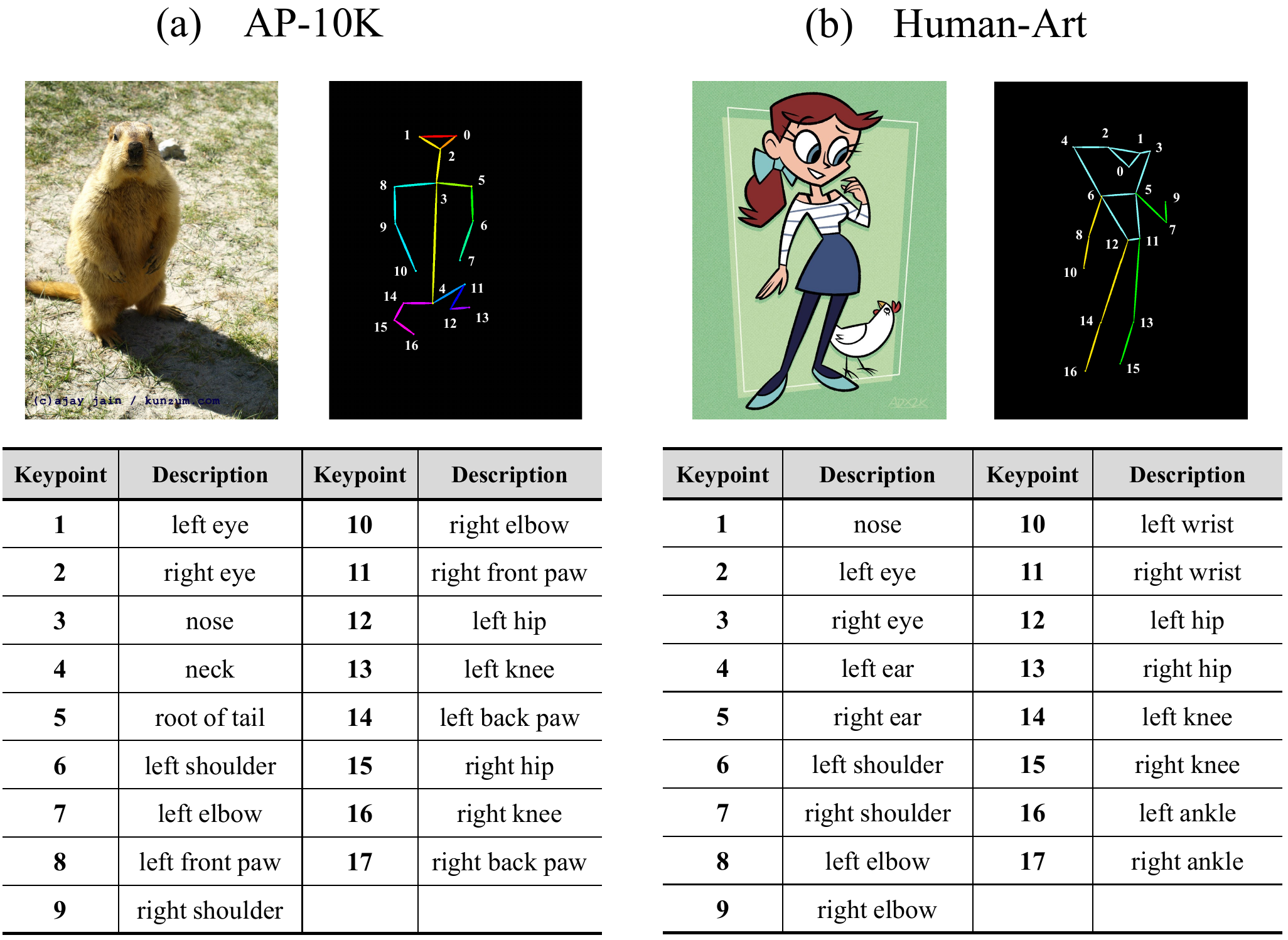}
            \caption{The definition (or description) of each keypoint on the AP-10K and Human-Art dataset. }
            \label{fig:app-kpt-definition}
        \end{figure}

        \para{Definition of animal and human pose.} Fig.~\ref{fig:app-kpt-definition} presents the definition of pose on the AP-10K~\cite{yu2021ap} and Human-Art~\cite{ju2023humanart} datasets, including pre-defined keypoint descriptions and the topological skeletons. For AP-10K, we adopt the 17-keypoint definition of pose for mammals, which is provided by the dataset. For Human-Art, the dataset provides two kinds of pose definitions for the human, \ie, 17-keypoints and 21-keypoints. To keep close to the definitions on animal poses, we employ the 17-keypoint definition of human pose as listed in Fig.~\ref{fig:app-kpt-definition}. To visualize the animal and human pose in OpenPose style, we utilize the code provided by the popular \texttt{mmpose}~\cite{mmpose2020} repository.
        
        \para{Prompt Templates for AP-10K.} Since the AP-10K dataset does not provide the image captions, we utilize a series of prompt templates following common practice~\cite{gal2022textualinversion}. We designed 10 templates as presented in Tab.~\ref{tab:app-prompt-template}, where the species name and background category are utilized to construct the textual prompt for each image. The AP-10K dataset predefines 54 species of 23 animal families and 8 background types~\cite{yu2021ap}, including {\it grass or savanna, forest or shrub, mud or rock, snowfield, zoo or human habitation, swamp or rivderside, desert or gobi, and mugshot}. When training, we randomly select one as the image caption. 

        \begin{table}[t]
            \caption{Prompt templates for the AP-10K dataset, where $\texttt{<CLS>}$ and $\texttt{<BG>}$ denote specie names and background types.}
            \centering
            \scalebox{0.80}{
            \begin{tabular}{c|l}
                \toprule
                ID & Prompt Templates \\
                \midrule
                1 & \texttt{A good photo of <CLS>.} \\
                2 & \texttt{A photo of <CLS> in the <BG>.} \\
                3 & \texttt{There is <CLS> on the <BG>} \\
                4 & \texttt{There are some <CLS> lying in the <BG>.} \\
                5 & \texttt{Some <CLS> are in the <BG>.} \\
                6 & \texttt{A close photo of <CLS>.} \\
                7 & \texttt{In the <BG>, there are several <CLS>.} \\
                8 & \texttt{This is a clear photo of <CLS> in the <BG>.} \\
                9 & \texttt{Several <CLS> are on the <BG>.} \\
                10 & \texttt{A <CLS> stands on the <BG>.} \\
                \bottomrule
            \end{tabular}}
            \label{tab:app-prompt-template}
        \end{table}
    
    \section{More Discussions}
    \label{sec:app-discussion}

        \begin{figure}[t]
            \centering
            \includegraphics[width=0.96\linewidth]{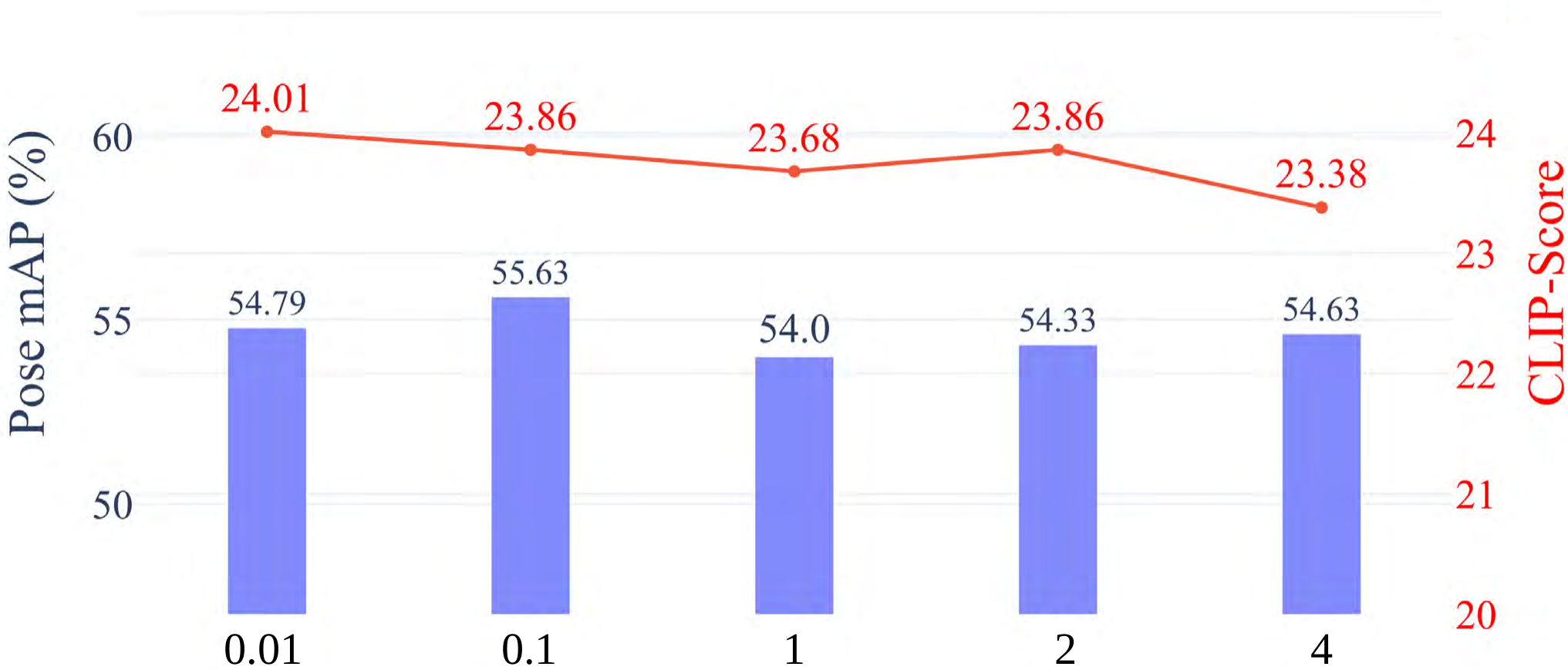}
            \caption{Discussions of the loss coefficient $\eta$.}
            \label{fig:app-ablation-loss-scale}
        \end{figure}

        \para{Discussions of the loss coefficient $\eta$.} We search for the optimal setting of the loss weighting $\eta$ for our proposed heatmap loss $\mathcal{L}_{ht}$ as reported in Fig.~\ref{fig:app-ablation-loss-scale}. While too large $\eta$ decreases the CLIP-Score and pose mAP, we set $\eta = 0.1$ in our experiments, which achieves the optimal pose mAP of $55.63\%$ and competitive CLIP-Score of $23.86$. 
        

        \begin{figure}[t]
            \centering
            \includegraphics[width=0.96\linewidth]{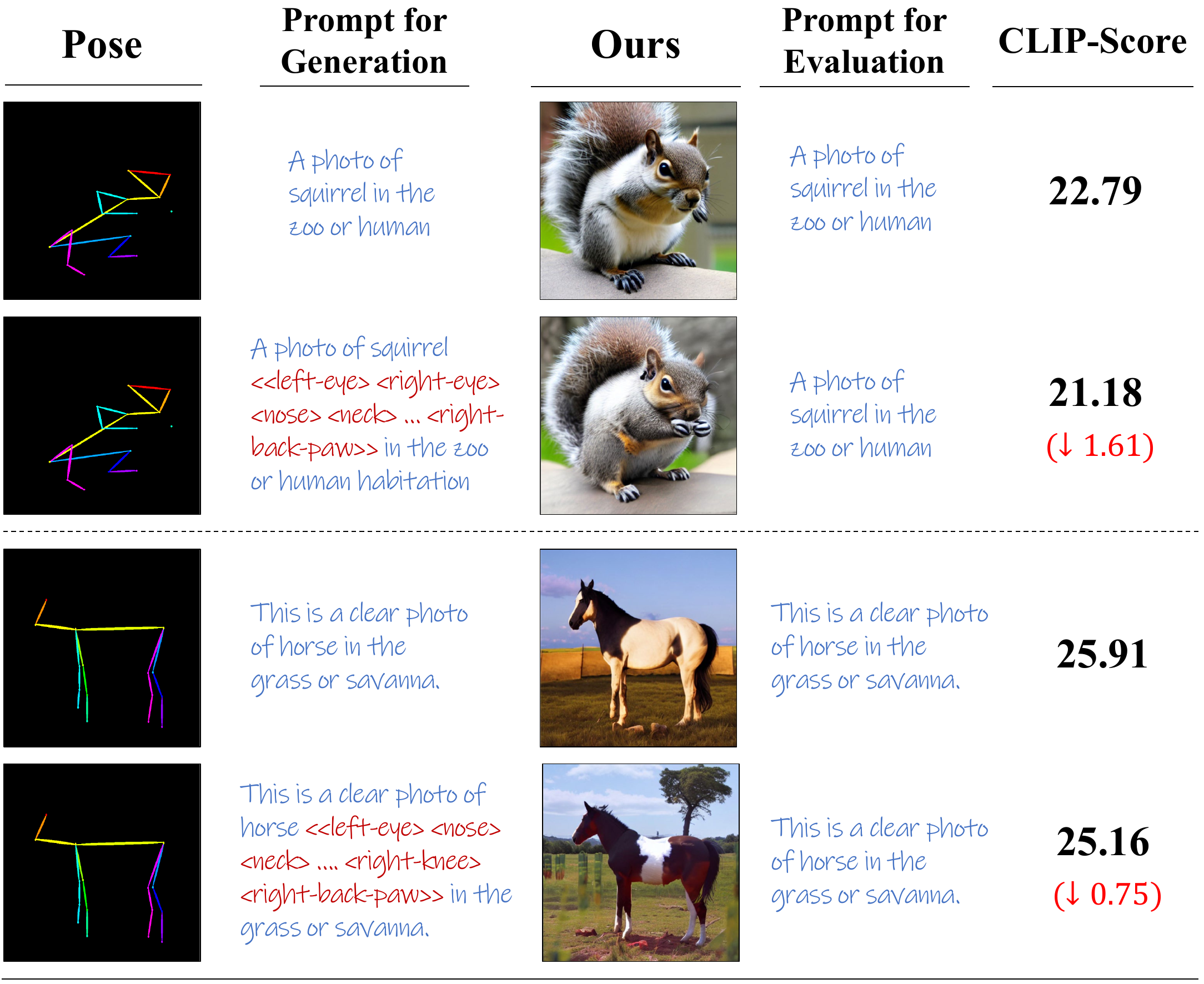}
            \caption{An illustration on the decrease of CLIP-Score due to inference-evaluation discrepancy.}
            \label{fig:app-clipscore-analysis}
        \end{figure}

        \para{Analysis of the CLIP-Score.} Here we provide some examples to explain the decrease of CLIP-Score as illustrated in the main paper, \S{\bf 4.4} Line 427. Since the vocabulary of the pretrained CLIP model does not contain the newly introduced keypoint tokens of $\mathcal{V}_{kpt}$ in the main paper \S{\bf 3.2}, we remove such tokens to compute the CLIP-Score for comparison. This operation causes marginal decrement of CLIP-Score as shown in Fig.~\ref{fig:app-clipscore-analysis}, where the CLIP-Score reduces by $1.61$ and $0.75$. This is because of the inference-evaluation discrepancies. Despite the visual difference between the images generated with and without the new keypoint tokens, our method still shows high text-image alignments. Therefore, we think the minimal decrease of CLIP-Score is acceptable. 

    \section{More Visualized Results}
    \label{sec:app-more-results}
         We provide additional qualitative results to supplement the main paper. Details are as follows. 
    
        \para{Full visualized results of cross-attention maps.} We show all the cross-attention maps of the vanilla ControlNet~\cite{zhang2023adding} and our {\it SP-Ctrl} method in Fig.~\ref{fig:app-controlnet-cross-attention-maps} and Fig.~\ref{fig:app-spctrl-cross-attention-maps}, where the time step here denotes the steps of adding noises. These attention maps are averaged across different transformer blocks at each time step. As shown in the figure, compared with the baseline ControlNet, the keypoint tokens attend to the positions of each keypoint more accurately. Though we only compute the $\mathcal{L}_{ht}$ among the $3^{rd}$ transformer blocks during the $250$$\sim$$500$ time steps following the best practice, we notice that the cross-attention maps after the $250$$\sim$$500$ time steps also attend to the keypoint positions. This fact indicates that constraining the cross-attention maps during the $250$$\sim$$500$ time steps implicitly regularizes the attention maps at other time steps, all contributing to the learning of new keypoint tokens. 
        
        \para{More visualized comparisons between other popular methods and ours.} Here we present more examples of pose-guided image synthesis on the AP-10K and Human-Art dataset in Fig.~\ref{fig:app-comparison-ap10k} and Fig.~\ref{fig:app-comparison-humanart} to show the effectiveness of our SP-Ctrl. As shown in the figure, while ControlNet and other methods might fail to interpret certain keypoints, such as the limbs, our method shows advantages in aligning with the detailed poses on both animal- and human-centric generation tasks. These results demonstrate the effectiveness of our method in pose controllable generation with sparse signals.  

        \para{More visualized comparisons on different pose guidance.} Fig.~\ref{fig:app-different-condition} presents more visualized examples of ControlNet with different conditions and ours. The masks fail to control the keypoint positions. While the depth map achieves precise control over pose, it also constrains the shape of generated animals. The OpenPose signal provides pose guidance for ControlNet but may fail when meeting complex poses or overlapped local structures. In contrast, our method achieves better control over sparse pose signals.

        \para{More examples to show the advantages of our method.} Benefiting from the precise pose control under the sparse pose guidance, our SP-Ctrl shows several appealing properties for applications. Compared to dense signals like depth, our method exhibits more diverse results in object shapes, as shown in Fig.~\ref{fig:app-diverse-shape}. Moreover, due to the category-agnostic characteristics of sparse pose, our method enables cross-species generation. As shown in Fig.~\ref{fig:app-cross-species}, despite the discrepancies in the action and skeleton proportions among different animal species, our method can produce promising results of different animals sharing the same pose. Additionally, since the sparse pose signals do not necessarily rely on the pretrained pose estimators, it enjoys great flexibility in pose editing and creation. Fig.~\ref{fig:app-editing} showcases several examples. Such results show the great potential of sparse pose signals in spatially controllable generation.

    \section{Limitations and Future Work}
    \label{sec:app-limitations}
        \begin{figure}[t]
            \centering
            \includegraphics[width=\linewidth]{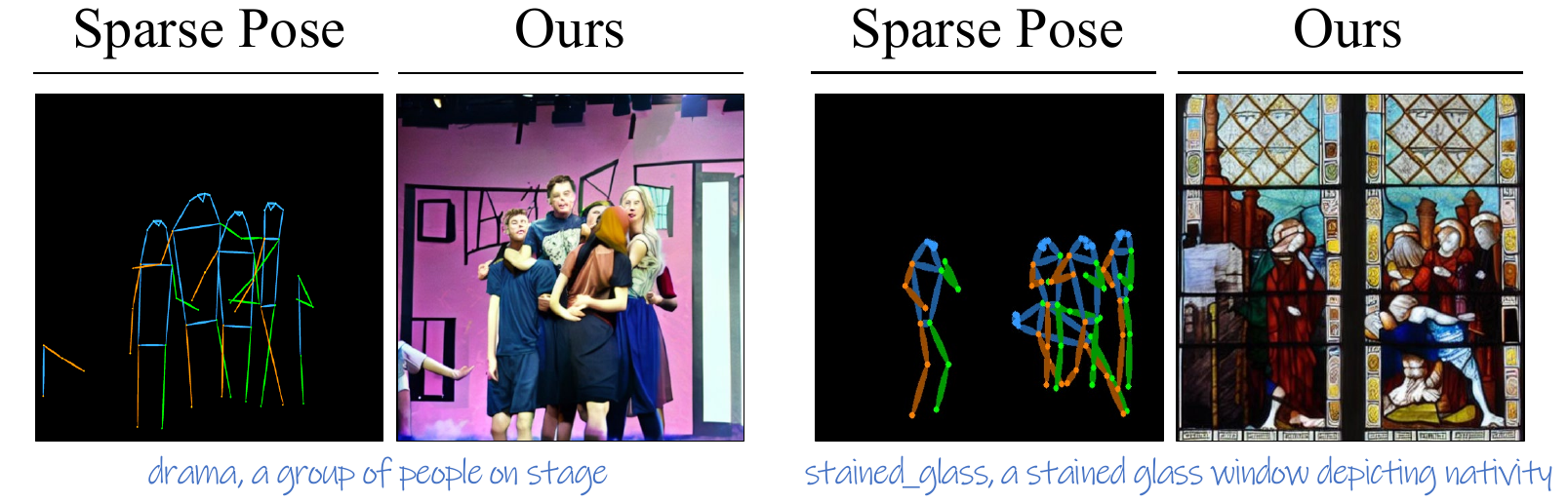}
            \caption{Examples of multiple-instance generation with pose signals, which is a more challenging case with overlaps, occlusion, and interactions of subjects.}
            \label{fig:app-limitation}
        \end{figure}
    
        By far, we have demonstrated the effectiveness of our SP-Ctrl in pose-guided text-to-image generation using sparse signals, which achieves performance nearly comparable to dense signal-based methods in terms of pose alignment. However, a significant gap in pose accuracy ($>25\%$) remains between synthesized images and real ones, particularly for rare or complex poses. One possible solution is to leverage synthesized images to augment the pose diversity, particularly for complex ones, to enhance the perception and pose-alignment of generative models. Another crucial challenge is pose-controllable generation involving multiple instances. Although our method has shown promising results in Fig.~\ref{fig:app-comparison-ap10k} and Fig.~\ref{fig:app-comparison-humanart}, further researches are required to address the multiple-instance generation with pose signals, which is a more challenging case with overlaps, occlusions, and interactions between subjects as presented in Fig.~\ref{fig:app-limitation}. We leave this for future work.

        \begin{figure*}[b]
            \centering
            \includegraphics[width=0.85\linewidth]{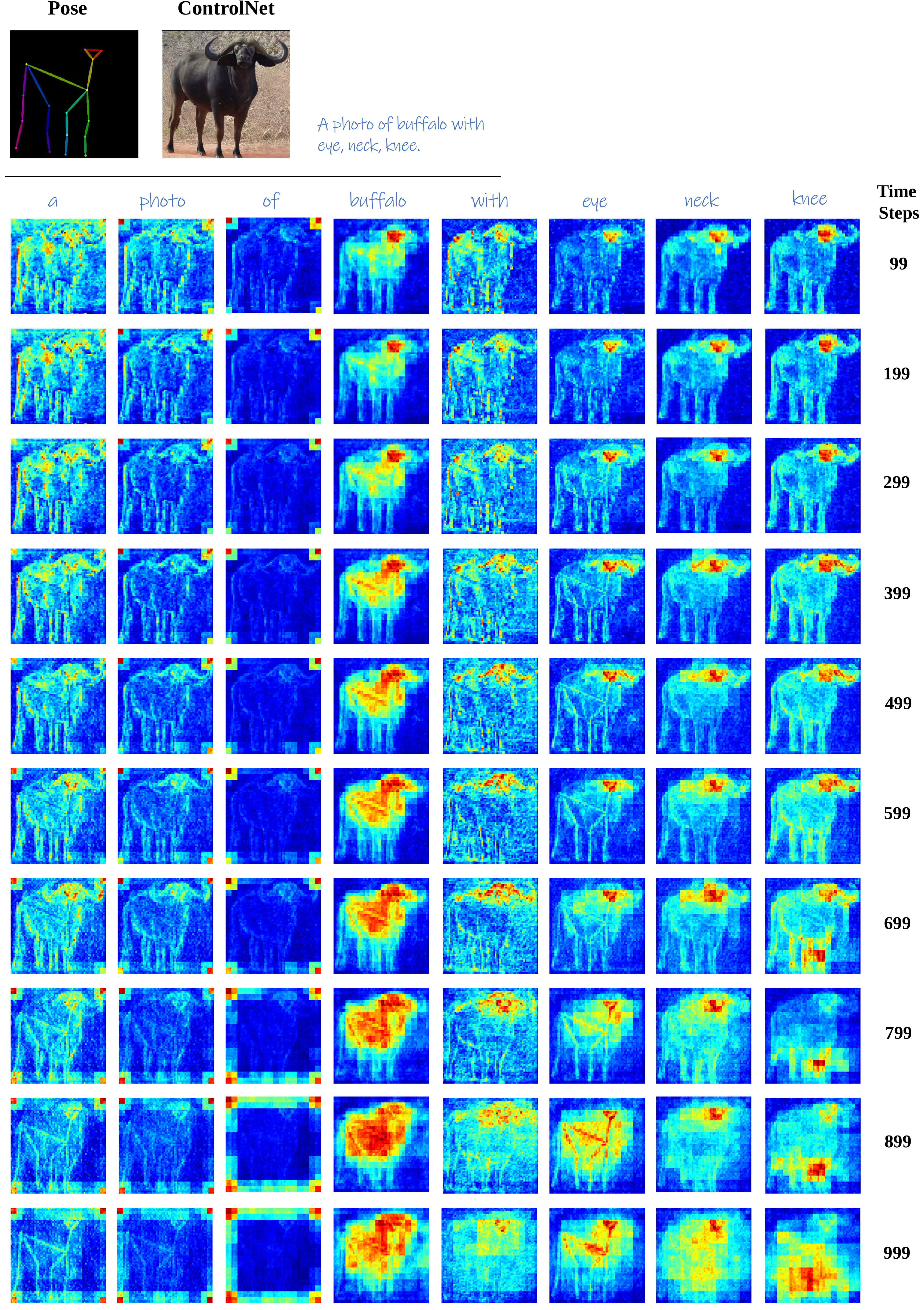}
            \caption{Visualized cross-attention maps of ControlNet at different time steps, which are averaged from all cross-attention layers. The time step here denotes the steps of adding noises.}
            \label{fig:app-controlnet-cross-attention-maps}
        \end{figure*}

        \begin{figure*}[b]
            \centering
            \includegraphics[width=1.0\linewidth]{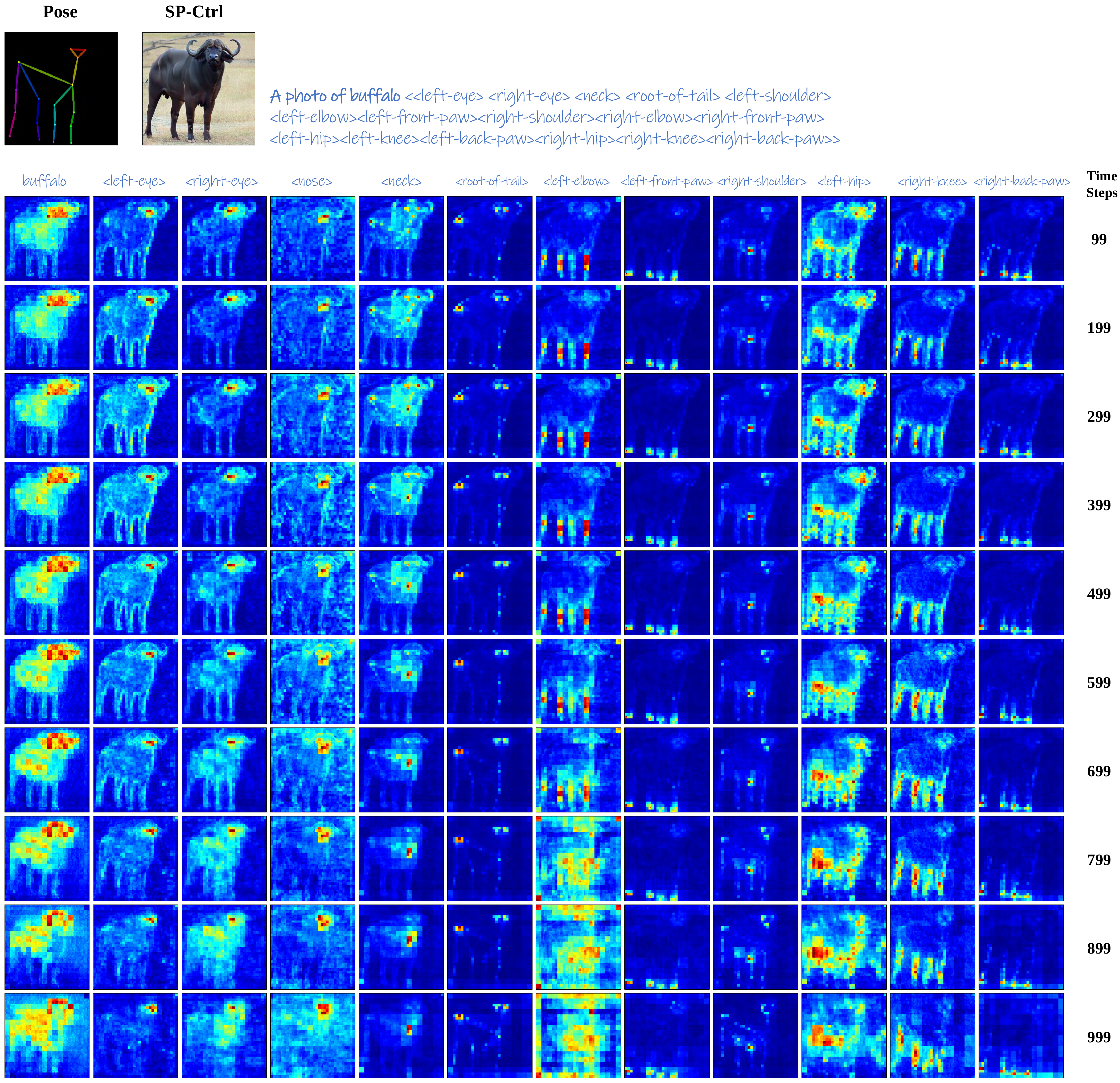}
            \caption{Visualized cross-attention maps of our {\it SP-Ctrl} at different time steps, which are averaged from all cross-attention layers. For page width limitations, we present the cross-attention maps of distinct keypoint for visualization. The time step here denotes the steps of adding noises. }
            \label{fig:app-spctrl-cross-attention-maps}
        \end{figure*}

        \begin{figure*}[t]
            \centering
            \includegraphics[width=0.98\linewidth]{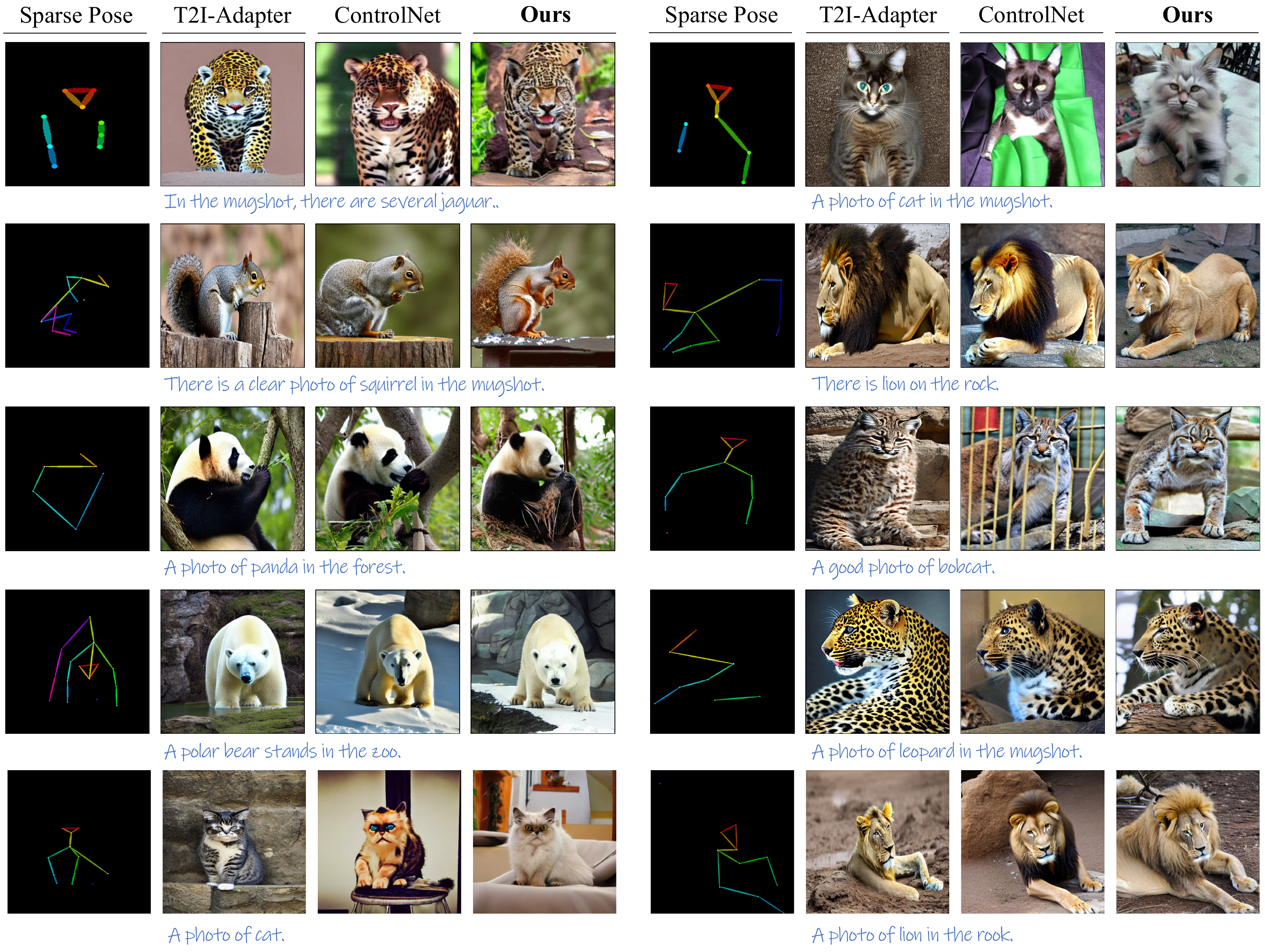}
            \caption{More visualized examples of our method and other popular methods on the AP-10K dataset.}
            \label{fig:app-comparison-ap10k}
        \end{figure*}

        \begin{figure*}[t]
            \centering
            \includegraphics[width=0.8\linewidth]{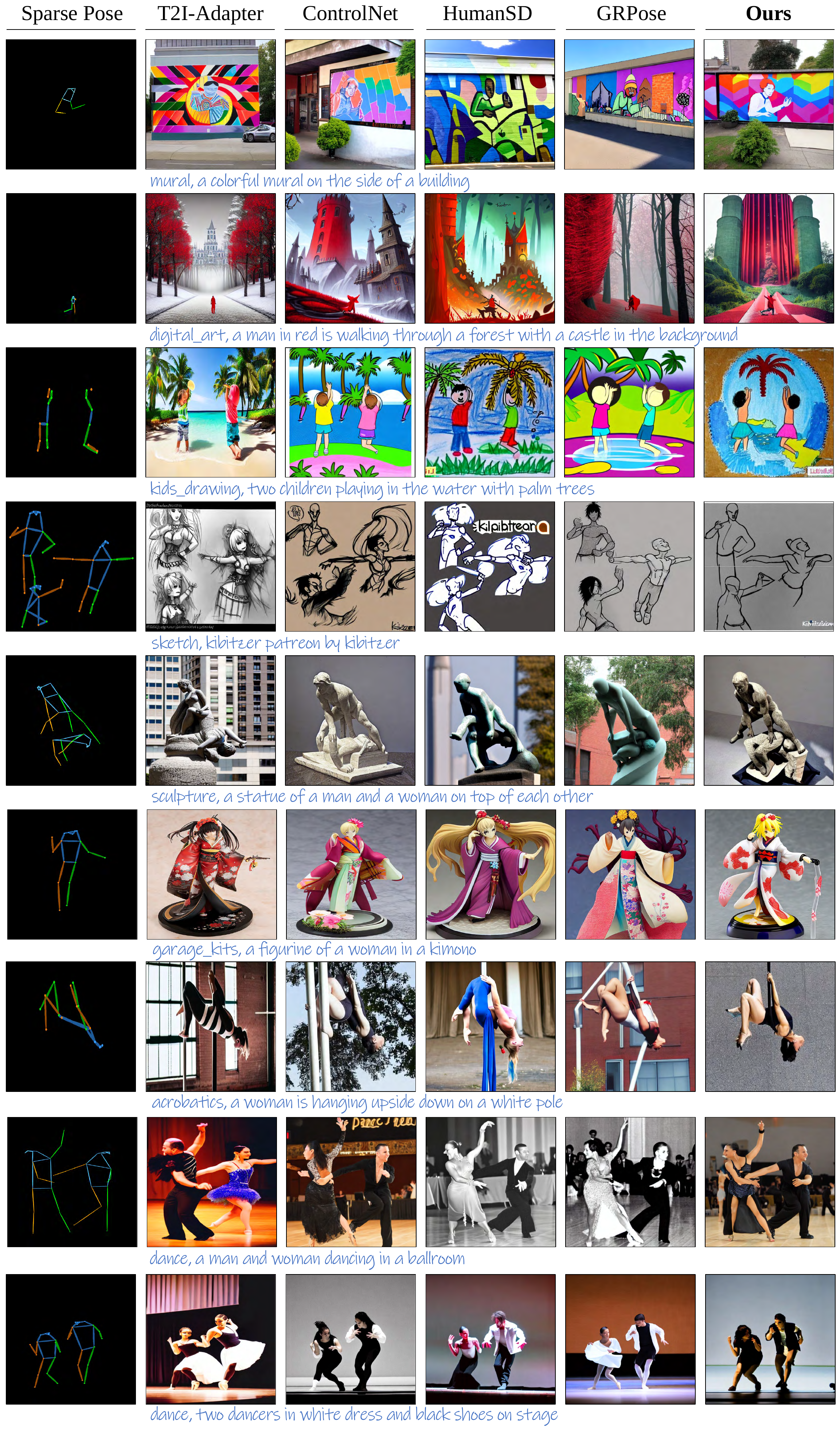}
            \caption{More visualized examples of our method and other popular methods on the Human-Art dataset.}
            \label{fig:app-comparison-humanart}
        \end{figure*}
        
        \begin{figure*}[t]
            \centering
            \includegraphics[width=0.78\linewidth]{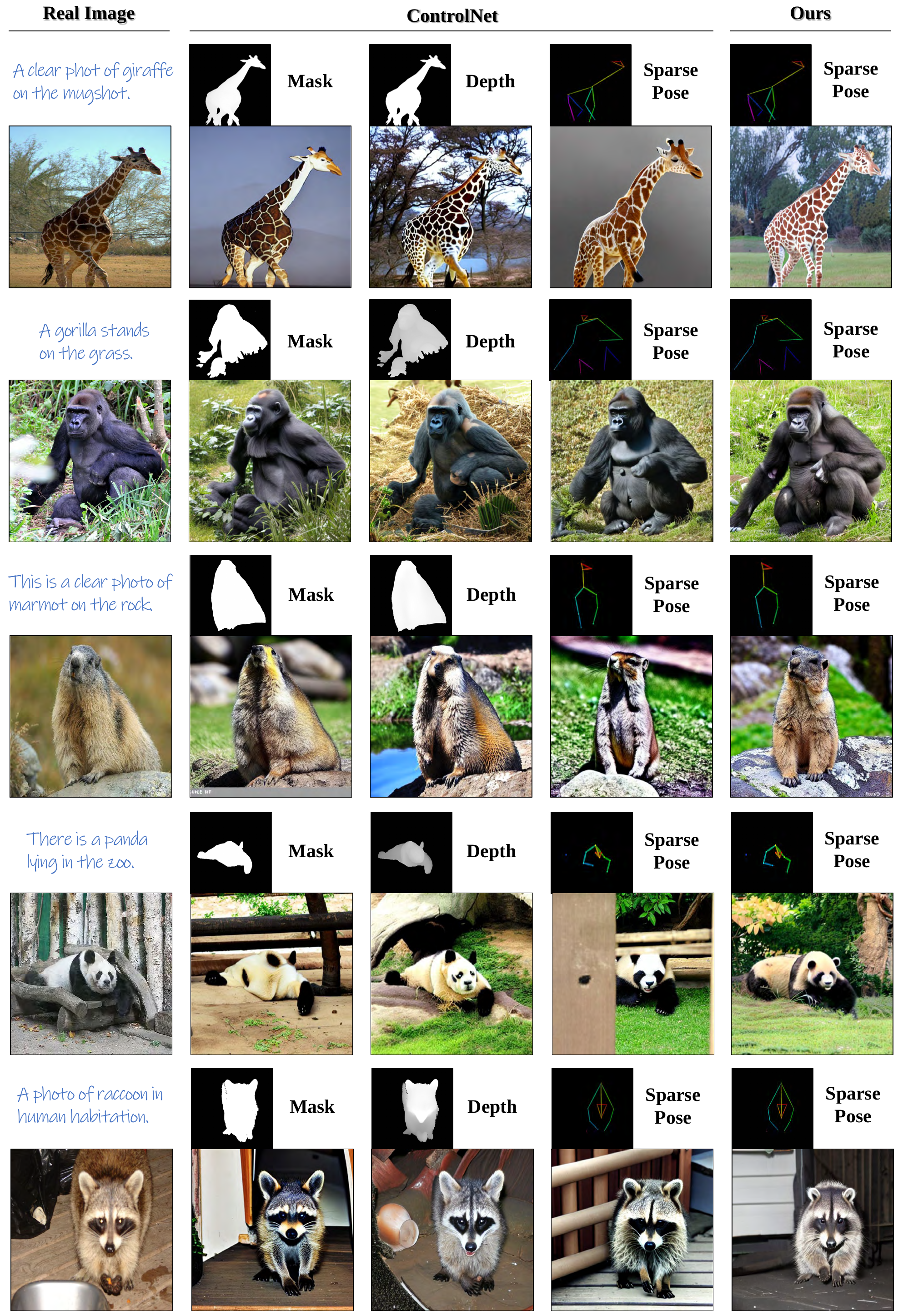}
            \caption{More visualized examples generated with different conditional guidance of ControlNet and our method.}
            \label{fig:app-different-condition}
        \end{figure*}

        \begin{figure*}[t]
            \centering
            \includegraphics[width=0.95\linewidth]{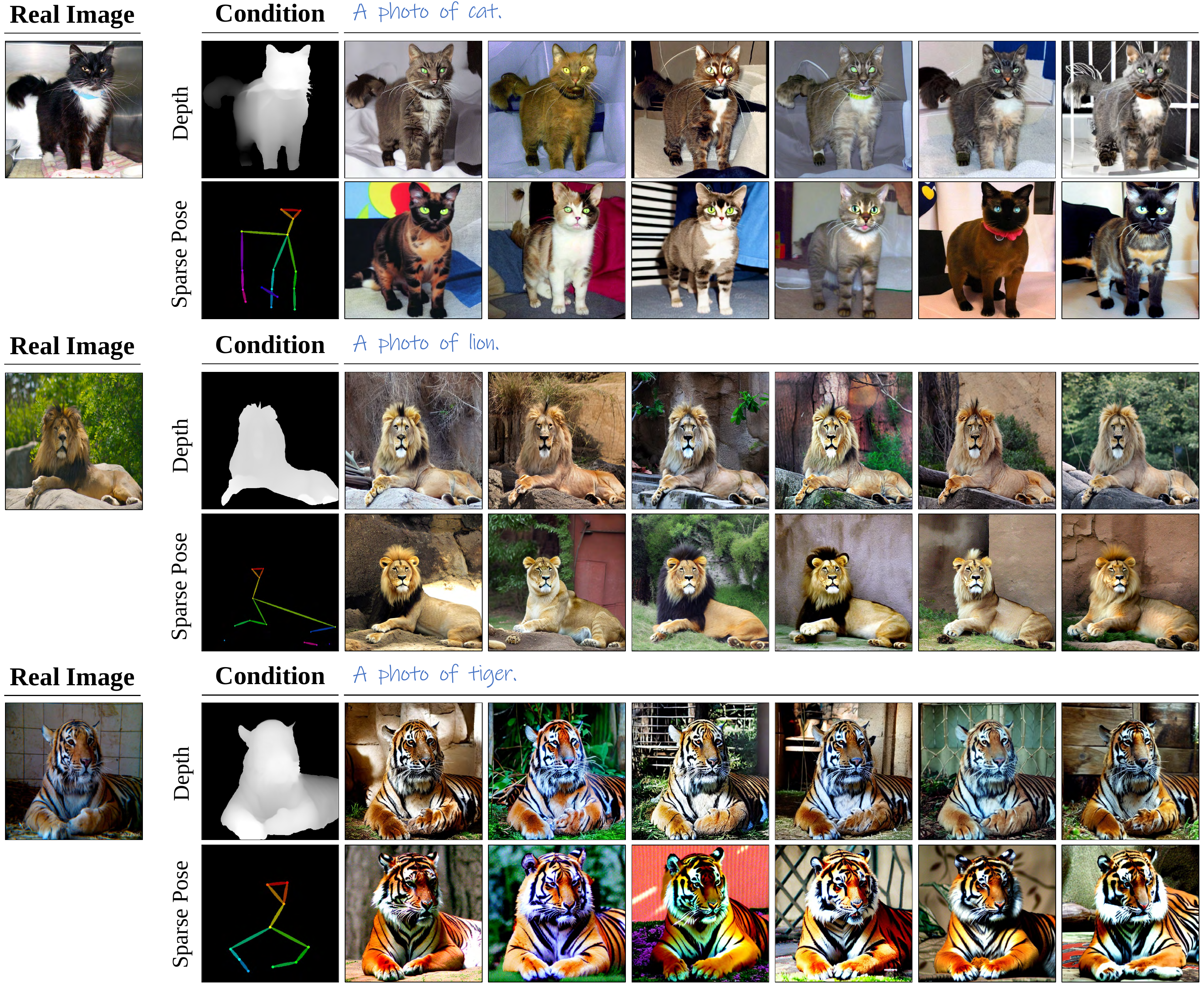}
            \caption{More visualized examples to showcase the shape diversity of synthesized images. When achieving precise pose control comparable to depth maps, our method can generate more diverse results with sparse signals, especially in shapes and contours. }
            \label{fig:app-diverse-shape}
        \end{figure*}

        \begin{figure*}[t]
            \centering
            \includegraphics[width=0.9\linewidth]{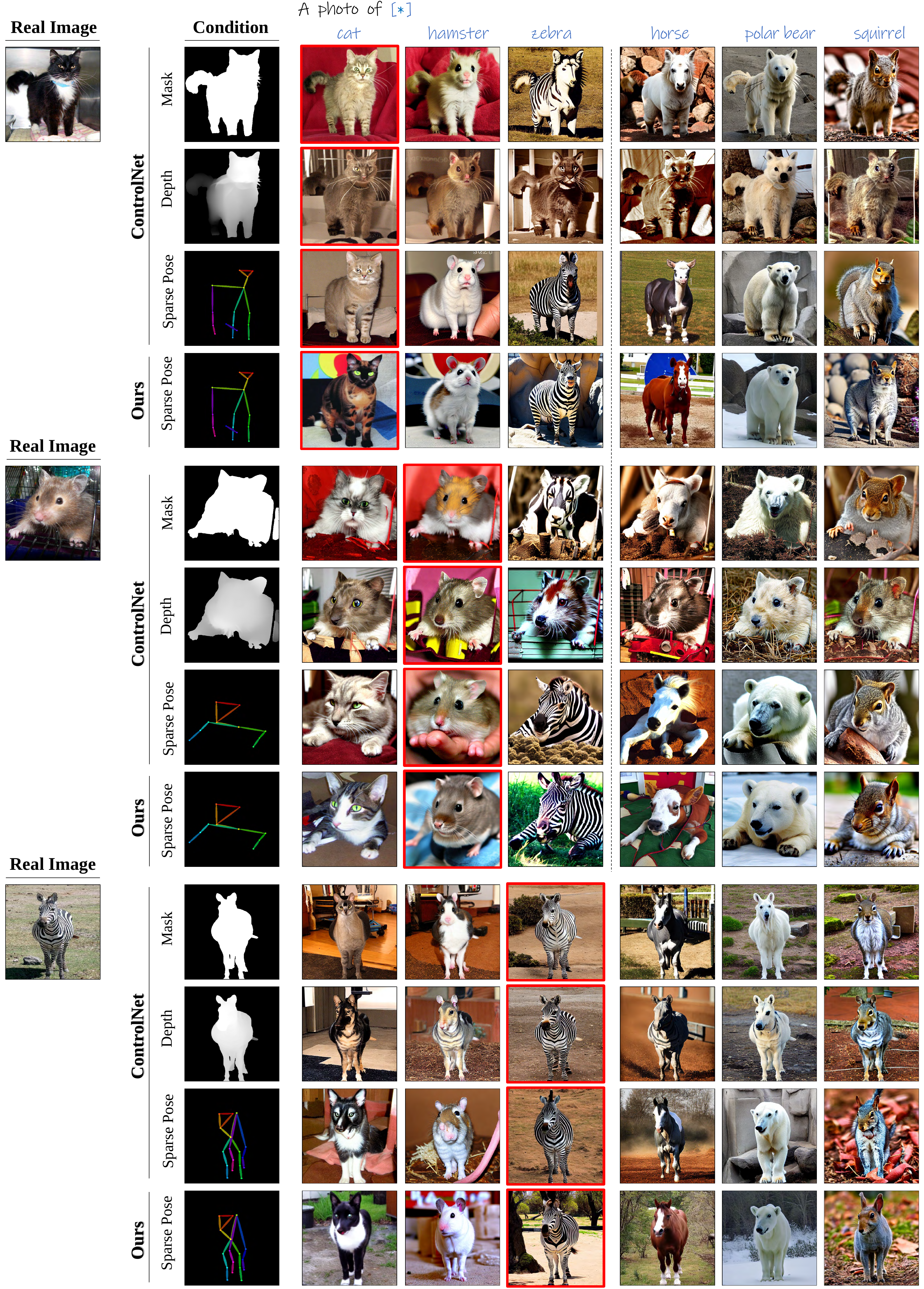}
            \caption{More visualized examples to showcase the cross-species generation ability. Our method can not only generate other species of animals but also keep high pose accuracy and image fidelity. }
            \label{fig:app-cross-species}
        \end{figure*}

        \begin{figure*}[t]
            \centering
            \includegraphics[width=\linewidth]{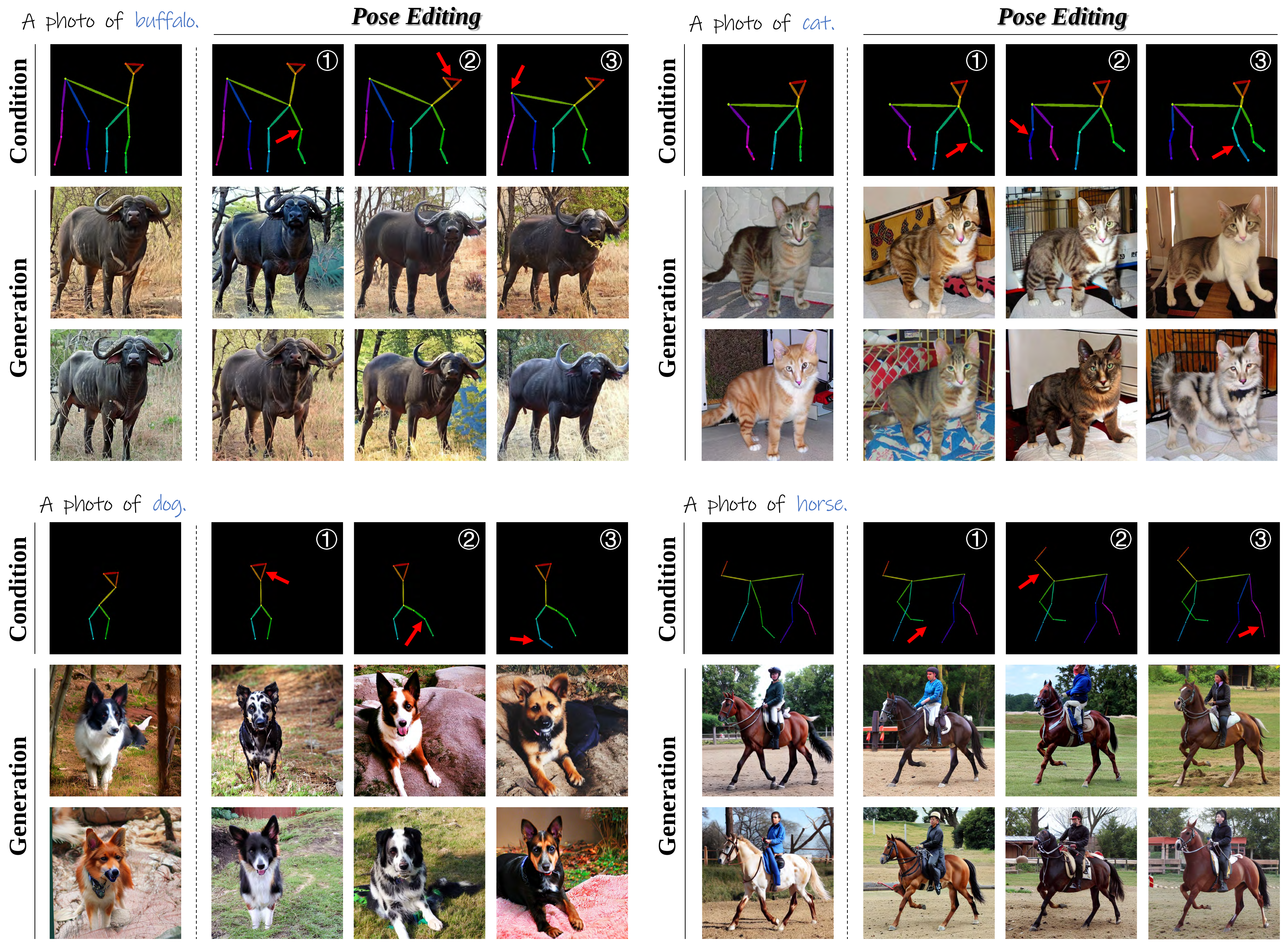}
            \caption{More visualized examples to showcase the generated image with edited pose signals, where sparse poses can be easily edited by changing the positions of keypoints. }
            \label{fig:app-editing}
        \end{figure*}

\end{document}